\documentclass[11pt]{article}

\usepackage[preprint]{acl}

\hypersetup{
      pdftitle={How Much is a Human Right Worth? ECtHR-NPD: A Benchmark for Predicting Non-Pecuniary Damage Awards},
      pdfauthor={Yanyi Pu; Damian A. Gonzalez-Salzberg; Zheng Yuan; Nikolaos Aletras},
      pdfsubject={A benchmark for predicting non-pecuniary damage awards at the European Court of Human Rights},
      pdfkeywords={legal NLP, ECtHR, non-pecuniary damage, benchmark, regression},
      pdflang={en-GB},
      pdfdisplaydoctitle=true
    }

\usepackage{times}
\usepackage{latexsym}
\usepackage{booktabs}
\usepackage{graphicx}
\usepackage{tabularx}
\usepackage{tikz}
\usetikzlibrary{positioning,arrows.meta}
\usepackage{textcomp}
\newcommand{\EUR}[1]{\texteuro{}#1}
\usepackage{array}
\usepackage{algorithm}
\usepackage{algpseudocode}
\usepackage{url}
\usepackage{amssymb}
\usepackage{float}
\usepackage{amsmath}
\algrenewcommand\algorithmicrequire{\textbf{Input:}}
\algrenewcommand\algorithmicensure{\textbf{Output:}}

\usepackage[T1]{fontenc}

\usepackage[utf8]{inputenc}

\usepackage{microtype}

\usepackage{inconsolata}

\usepackage{enumitem}
\usepackage{rotating}

\makeatletter
\renewcommand{\paragraph}{%
  \@startsection{paragraph}{4}%
  {\z@}{2pt \@plus 1pt \@minus 1pt}{-1em}%
  {\normalfont\normalsize\bfseries}%
}
\makeatother

\title{How Much is a Human Right Worth? ECtHR-NPD: A Benchmark for Predicting Non-Pecuniary Damage Awards}

\author{
  \textbf{Yanyi Pu\textsuperscript{$\sigma$}} \quad
  \textbf{Damian A. Gonzalez-Salzberg\textsuperscript{$\beta$}}\\
  \textbf{Zheng Yuan\textsuperscript{$\sigma$}} \quad
  \textbf{Nikolaos Aletras\textsuperscript{$\sigma$}}
\\
  \textsuperscript{$\sigma$}School of Computer Science, University of Sheffield \\
  \textsuperscript{$\beta$}Birmingham Law School, University of Birmingham \\
    \texttt{\{ypu17,zheng.yuan1,n.aletras\}@sheffield.ac.uk}\\
  \texttt{d.a.gonzalezsalzberg@bham.ac.uk}
}

\begin{document}
\maketitle

\begin{abstract}

Existing legal benchmarks cover diverse tasks, while continuous \textit{monetary remedies} remain comparatively underexplored. We introduce \textbf{ECtHR-NPD}, to the best of our knowledge, the first benchmark for predicting non-pecuniary damage (NPD) awards at the European Court of Human Rights (ECtHR) from case information when no statutory formula or explicit calculation rule determines the amount. ECtHR-NPD contains 14,575 cases with case-level awards in nominal euros, chronological splits, and a protocol separating target construction from model input. We evaluate a battery of methods, including constant predictors, gradient-boosted trees, retrieval methods, fine-tuned encoder language models (LMs), prompted decoder LMs, and knowledge-augmented agents. Our results show that more sophisticated LM and agentic approaches do not consistently outperform the strongest feature-based baseline. All model families struggle to identify zero awards and to calibrate high-award predictions, with further degradation on the Challenging test view, making ECtHR-NPD a challenging testbed for current state-of-the-art open-weight and proprietary LMs.\footnote{Data is available on \href{https://huggingface.co/datasets/YanyiPU716/ECtHR-NPD}{Hugging Face}. Code and documentation are available on \href{https://github.com/YanyiPU/ECtHR-NPD}{GitHub}.} 

\end{abstract}

\section{Introduction}

\begin{figure}[!t]
  \centering
  \includegraphics[width=1\columnwidth]{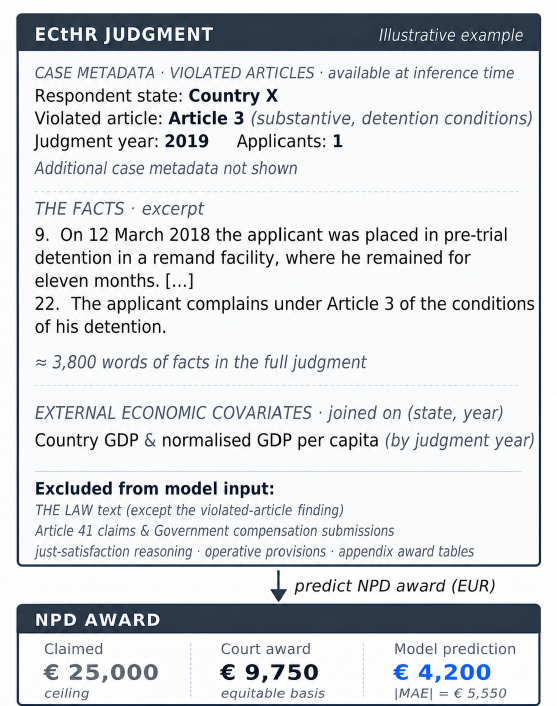}
    \caption{The ECtHR-NPD task. Models receive  
    case metadata, violated Convention articles, case facts, and external macroeconomic covariates. Award-related material is excluded from model input. The model outputs the case-level non-pecuniary damage award. 
    }
  \label{fig:task}
\end{figure}

Legal NLP benchmarks typically cover classification, retrieval, extraction, generation, and legal reasoning tasks~\citep{chalkidis2022lexglue,niklaus2023lextreme,guha2023legalbench,fan2026lexam,shi2026plawbench}, but few evaluate continuous monetary outcomes. Existing numerical legal tasks often concern outcomes bounded by statutory ranges or determined by explicit rules, including sentencing terms~\citep{xiao2018cail,bi2023numljp} or tax liabilities~\citep{holzenberger2020dataset}. However, courts may need to make quantitative decisions, such as determining the amount of a monetary remedy following a legal violation, without a published or known formula. 

The ECtHR is an international court that hears applications alleging that a member state has breached human rights protected by the European Convention on Human Rights (ECHR), covering rights such as the right to life, liberty, and fair trial. When the Court finds a Convention violation, Article~41 allows it to award \textit{just satisfaction}, including non-pecuniary damage (NPD) for suffering, humiliation, and other types of intangible harm~\citep{schabas2015article41,echr2022justsatisfaction}. The Court assesses NPD on an \textit{equitable basis} rather than by precise calculation, and the recorded awards provide observable references to the Court's practice. Legal scholarship describes these awards as case-specific, empirically patterned, and only partly transparent~\citep{ichim2015just,altwicker2016measuring,fikfak2018changing,fikfak2020nonpecuniary,gonzalezsalzberg2021nonpecuniary}. Appendix~\ref{app:legal-background} gives further ECtHR and Article~41 background.

In this paper, we introduce \textbf{ECtHR-NPD}, a benchmark for predicting Article~41 NPD at the ECtHR. Figure~\ref{fig:task} provides an overview of the task, including the model inputs, withheld award-related information, and case-level prediction target.  \textbf{ECtHR-NPD} contains 14{,}575 judgments with validated case-level targets in nominal euros, two model input representations, chronological train/validation/test splits, and three diagnostic test views (ID/OOD/Challenging) (Section~\ref{sec:dataset}). The task is challenging because awards are zero-inflated and heavy-tailed, predictive signals are dispersed across long and heterogeneous judgments, and the relationship between case facts and award amounts is not given by an explicit calculation rule.

We evaluate six method families: constant predictors, gradient-boosted trees, retrieval methods, fine-tuned encoder LMs, prompted decoder LMs, and knowledge-augmented agents. \textsc{ECtHR-NPD} remains challenging across this broad range of approaches. The best test mean absolute error (MAE) is 10.2\% below the training-set median predictor. Prompted decoder LMs and agents are unstable and often fail to surpass simple statistical and feature-based baselines. All method families fail at identifying zero-award cases, and degrade sharply on both high-award cases and legally challenging cases.  Our experiments show that our benchmark exposes failure modes that existing classification-oriented legal NLP benchmarks do not measure. We make three contributions:

\begin{enumerate}[leftmargin=2em, topsep=2pt, itemsep=0pt]
  \item We introduce continuous non-pecuniary damage award prediction as a regression task, extending legal NLP evaluation to discretionary monetary remedies.
  \item We release \textbf{ECtHR-NPD}, a benchmark with validated targets, chronological splits, diagnostic test views, and structured annotations.
  \item We compare a broad set of methods and provide diagnostic evaluation across award ranges, test views, respondent states, and Convention articles.
\end{enumerate}

\section{Related Work}

\paragraph{Legal benchmarks.}
Legal NLP benchmarks cover increasingly diverse jurisdictions, but their supervised targets remain largely categorical, retrieval-based, textual, or rule-bounded. Prior ECtHR work has largely focused on violation prediction, rationale extraction, vulnerability classification, prior-case retrieval, selective prediction, and judicial disagreement~\citep{aletras2016predicting,chalkidis2019neural,chalkidis2021paragraph,xu2023vechr,santosh2024ecthrpcr,tyss2024selective,xu2024splitvote}.

Broader European resources include ECHR-OD, Swiss and UK court corpora, multilingual legal corpora, and aggregate benchmarks such as LexGLUE, LEXTREME, and LEXam~\citep{quemy2022echrod,niklaus2021swiss,ostling2023cambridge,chalkidis2023lexfiles,niklaus2024multilegalpile,chalkidis2022lexglue,niklaus2023lextreme,fan2026lexam}. Outside Europe, major benchmarks cover US legal reasoning, holding identification, and contract review~\citep{guha2023legalbench,zheng2021casehold,hendrycks2021cuad,henderson2022pileoflaw}, Chinese legal judgment prediction, legal question answering (QA), event detection, and LM evaluation~\citep{xiao2018cail,zhong2018legal,liu2023mlljp,gan2022numericalevidence,fei2024lawbench,zhong2020jecqa,yao2022leven}, and Indian judgment prediction, explanation, statute prediction, and multilingual evaluation~\citep{malik2021ildc,kapoor2022hldc,nigam2024predex,joshi2024iltur,vats2023llms,nigam2024nyayaanumana}. Recent work also evaluates LLMs across realistic legal practice scenarios using expert-designed rubrics~\citep{shi2026plawbench}. These benchmarks have substantially advanced legal NLP, but they do not evaluate discretionary monetary remedies as continuous targets.

\paragraph{Numerical prediction in the legal domain.}
Numerical legal prediction has mainly been studied in settings where numerical outcomes are constrained by legal rules or ranges. In criminal law, models are commonly used to predict sentencing terms jointly with charges and law articles~\citep{xiao2018cail,zhong2018legal,yang2019legal,liu2023mlljp}, or use numerical evidence to distinguish related charges~\citep{gan2022numericalevidence}. In tax law, SARA includes questions in which numerical tax liabilities are derived from statutory rules~\citep{holzenberger2020dataset}.

 A smaller body of work studies monetary outcomes determined with greater judicial discretion, including Brazilian airline-consumer immaterial damages~\citep{dalpont2023immaterial}, Taiwanese fatal-accident mental-suffering damages~\citep{hsieh2021legal}, and US jury-verdict valuation~\citep{conrad2017scenario}. These studies differ from ECtHR-NPD in jurisdiction, claim type, available evidence, target construction, and scale. Empirical legal scholarship has analysed human rights damages at both the ECtHR~\citep{altwicker2016measuring,fikfak2018changing,fikfak2020nonpecuniary} and the Inter-American Court~\citep{gonzalezsalzberg2021nonpecuniary}, providing motivation for predicting NPD awards. 

\section{The ECtHR-NPD Benchmark}
\label{sec:dataset}

\begin{figure}[t]
  \centering
  \includegraphics[width=\columnwidth]{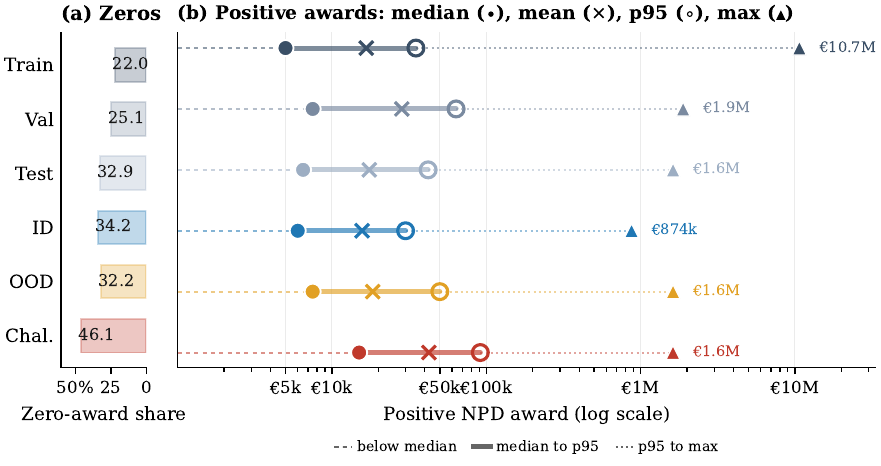}
  \caption{Award distribution across splits and diagnostic views. Panel~(a) shows the zero-award share. Panel~(b) shows the distribution of positive awards on a log scale, with markers for the median, mean, p95, and maximum.}
  \label{fig:award-dist}
\end{figure}

\subsection{Task Definition}
\label{sec:task-definition}

Given an ECtHR judgment in which the Court has found at least one Convention violation, the task is to predict the case-level Article~41 non-pecuniary damage (NPD) award. The model input $X$ contains case metadata, the \textsc{Facts} section, violated articles, and external macroeconomic covariates. The output $y \in \mathbb{R}_{\geq 0}$ is a nominal euro amount recorded in the judgment, including valid zero-award outcomes.  A model predicts \(\hat{y} \in \mathbb{R}_{\geq 0}\) from \(X\).

\subsection{Data Collection}
\label{sec:data-collection}

We collect English-language judgments from the Court's public HUDOC database~\citep{echr2026hudoc} using the March 2026 snapshot. We retrieve an initial pool of 18{,}367 English judgments and follow the established ECtHR document-processing setup used in prior ECtHR NLP resources~\citep{quemy2022echrod,chalkidis2019neural}. We then segment each judgment into its main structural components  (e.g., Procedure, The Facts, The Law, Article~41, and Operative Provisions). Appendix~\ref{app:structure} describes the common structure of ECtHR judgments. Figure~\ref{fig:task} shows which parts are used as model input and which parts are withheld for label construction. 

Models do not receive applicants' Article~41 claims, Government submissions on compensation, the Court's just-satisfaction reasoning, operative provisions, or appendix award tables. These materials are used only to construct and validate the NPD target. The full extraction and validation process is reported in Appendix~\ref{app:dataset-details}.

\subsection{Input Representation} 
\label{sec:constructing-input}
We use two forms of input in our experiments.

\paragraph{Raw-text input.}
We provide the model with: (i) case metadata, including respondent country, judgment year, court formation, procedural status, and related case descriptors; (ii) the \textsc{Facts} section; (iii) the Convention articles found to be violated; and (iv) external macroeconomic covariates joined by respondent state and judgment year. This setting tests whether models can identify information relevant to NPD prediction directly from the long factual description of the case.

\paragraph{Structured input.}
The structured input replaces the \textsc{Facts} section with a compact set of case features extracted by our pipeline (Appendix~\ref{app:dataset-details}), while retaining the metadata, violated Convention articles, and macroeconomic covariates. This setting reduces the context by making the information relevant to NPD prediction explicit. For prompted decoder LMs, these features are written as a short key-value text block. A simplified example is shown in Figure~\ref{fig:serialisation-example}. The tree models use the same feature groups in tabular form (Section~\ref{sec:baselines}). 

\begin{figure}[!t]
    \centering
    \fbox{%
        \begin{minipage}{0.90\columnwidth}
        \vspace{3pt}
        \small\ttfamily
        Respondent state: Armenia\\
        Judgment year: 2020\\
        Court formation: Chamber\\
        Violated Article: Article 5\\
        Violation type: substantive\\
        Applicants: 1\\
        Applicant age group: adult\\
        Applicant sex: male\\
        Violation duration: 18 months\\[2pt]
        \normalfont\itshape Additional structured features omitted.
        \vspace{3pt}
        \end{minipage}%
    }
    \caption{Abridged structured input from an ECtHR-NPD case.}
    \label{fig:serialisation-example}
\end{figure}

\subsection{Target Construction}
\label{sec:target-construction}

The target $y \in \mathbb{R}_{\geq 0}$ is the total case-level NPD amount in nominal euros, including valid zero-award outcomes. We extract candidate NPD amounts from the Article~41 section and validate them against operative provisions and appendix award tables where available. Candidate targets must pass checks for award-head separation, per-applicant sum consistency, currency normalisation, and recoverability from the operative provisions. Zero targets are retained only when supported by an accepted legal rationale described in Appendix~\ref{app:taxonomy}. Full validation contracts are reported in Appendix~\ref{app:target-construction}.

We use nominal euro amounts rather than adjusting them for inflation. Deflating the awards would require choices about an external price index, data source, and base year, whereas the nominal amounts remain directly traceable to the judgment. Judgment year and state-year macroeconomic covariates are provided as model inputs. 

\paragraph{Data Filtering.}  
We exclude cases with no recorded Convention violations, cases where the NPD amount is not recoverable, cases where the NPD amount is bundled inseparably with other claim heads (such as pecuniary damages or costs), cases lacking an accepted legal rationale for a zero-award outcome, and cases denominated in pre-euro currencies (historical Article~50 claims). 
These filters, developed by a legal expert, yield a final dataset of 14{,}575 cases with validated continuous euro targets. The full filtering and target validation procedure is detailed in Appendix~\ref{app:source-corpus}.

\begin{table}[!t]
\small
\setlength{\tabcolsep}{4pt}
\begin{tabular}{l r l l}
\toprule
Set & $n$ & Period / source & Mean TV \\
\midrule
\multicolumn{4}{l}{\textit{Chronological splits}} \\
Train & 10{,}217 & 1968--Jul.~2019 & -- \\
Validation & 1{,}461 & Jul.~2019--Dec.~2021 & -- \\
Test pool & 2{,}897 & Dec.~2021--Mar.~2026 & -- \\
\midrule
\multicolumn{4}{l}{\textit{Diagnostic test views}} \\
ID & 1{,}000 & Test pool & 0.025 \\
OOD & 1{,}897 & Test pool & 0.211 \\
Challenging & 699 & Test pool & 0.435 \\
\bottomrule
\end{tabular}
\caption{Chronological splits and diagnostic test views. Mean Total Variation (TV) measures distributional distance from the train plus validation
reference distribution across the eight matching dimensions; lower values
indicate closer alignment. \textit{Challenging} is an overlapping diagnostic subset.}
\label{tab:splits-views}
\end{table}

\subsection{Data Splits and Diagnostic Views}
\label{sec:splits-views}

We partition the dataset chronologically by judgment date into training (70\%), validation (10\%), and test (20\%), following prior work~\citep{chalkidis2019neural,chalkidis2022lexglue}. Same-day ties are broken deterministically by HUDOC item identifier, so no case appears in more than one split. This chronological split prevents temporal leakage by keeping later judgments out of the data used to fit and select models, while reserving the latest judgments for evaluating temporal generalisation. The resulting test pool comprises 2{,}897 judgments. Table~\ref{tab:splits-views} summarises the split sizes and diagnostic views.  Figure~\ref{fig:award-dist} shows the corresponding award distributions. 

We further partition the test pool into three test subsets:

\paragraph{In-Distribution (ID).} We use a greedy quota-matching algorithm (Algorithm~\ref{alg:greedy-matching}) to select (n = 1{,}000) from the test pool. This view aligns the ID subset with the empirical distribution of the training and validation references across predefined structural dimensions. Table~\ref{tab:matching-diagnostics} reports the matching dimensions, their weights, and the resulting distributional differences. The ID view evaluates model performance on temporal generalisation for cases that are structurally similar to the reference.

\paragraph{Out-of-Distribution (OOD).} The residual cases in the test set (n = 1{,}897) constitute the OOD view. These cases are less represented in the reference distribution along the matching dimensions. We use OOD to test whether models generalise beyond the structurally similar cases captured by the ID view.

\paragraph{Challenging.} We additionally identify an overlapping diagnostic subset (n = 699) using predefined criteria identified by a legal expert. It contains all Grand Chamber judgments, together with cases that have both multiple applicants and multiple concurrent Convention violations. The expert considered such cases particularly difficult to assess and quantify, and we use this subset to test whether they also pose greater difficulty for models.

\section{Experimental Setup}
\label{sec:experiments}

\subsection{Models}
\label{sec:baselines}

We evaluate six model families listed below. Full implementation details, including checkpoints, hyperparameters, preprocessing, retrieval settings, prompt templates, decoding parameters, and ReAct~\citep{yao2023react} tool policies, are reported in Appendix~\ref{app:experimental-design-setup}.

\paragraph{Constant baselines.}
We report the training-set median, training-set mean, and constant-zero predictors as weak baselines. The training-set median is the optimal constant predictor under MAE.

\paragraph{Gradient-boosted trees.}
We train CatBoost~\citep{prokhorenkova2018catboost}, XGBoost~\citep{chen2016xgboost}, and LightGBM~\citep{ke2017lightgbm} on the leakage-controlled structured
features rather than raw text (Appendix~\ref{app:trees}).

\paragraph{Retrieval baselines.}
We compare $k$-nearest-neighbour retrieval~\citep{cover1967nearest}, BM25~\citep{robertson2009probabilistic} implemented with BM25S~\citep{han2024bm25s}, and BGE-M3~\citep{chen2024bgem3}.
Prediction is the median award among retrieved training neighbours. Reference cases must precede the target judgment and share at least one violated Convention Article. These baselines test whether similarity to prior cases helps predict the award amount.

\paragraph{Encoder language models.}

We fine-tune ModernBERT~\citep{warner2025modernbert} and Legal-Longformer~\citep{chalkidis2023lexfiles} with a regression head on the raw-text input. A late-fusion variant concatenates the encoder \texttt{[CLS]} representation with a learned representation of the structured features.

\paragraph{Prompted decoder LMs.}

We evaluate Qwen3.5 variants (9B, 27B, and Plus)~\citep{qwen35blog}, GPT-OSS-20B~\citep{gptoss2025}, and GPT-5.4~\citep{openai2026gpt54} under zero-shot, fixed chain-of-thought (CoT)~\citep{wei2022cot}, and retrieved few-shot CoT prompting. CoT prompts were developed with a legal expert to guide the model through considerations relevant to the NPD assessment. Retrieved few-shot CoT added up to five prior training cases selected by similarity features. Their observed awards and Article~41 reasoning are included as references. Prompt templates were fixed before test evaluation. Full prompt design and retrieval settings are reported in Appendix~\ref{app:llm}.

\paragraph{Knowledge-augmented Agents.}
We evaluate a ReAct-style approach~\citep{yao2023react} with access to a structured legal knowledge base, train-only empirical priors, and eligible reference cases from the training set. Agent actions are
restricted to a whitelisted tool set, with no external web access. We run this setting on Qwen3.5-Plus and MiniMax-M2.7~\citep{minimax2026m27}. Full controller settings, tool definitions, access policies, and knowledge-base development are described in Appendix~\ref{app:react}.

\subsection{Evaluation Metrics}
\label{sec:metrics}
We use \textbf{mean absolute error (MAE)} in euros as our primary metric because it is an interpretable measure of average absolute monetary error per case. Since the target is zero-inflated and heavy-tailed, we also report complementary metrics for extreme-award sensitivity, typical and upper-tail error, rank association, magnitude calibration, and zero-award recognition (Table~\ref{tab:metrics}). Model selection and hyperparameter tuning use the validation split; the test set is reserved for final reporting.

\begin{table}[!t]
\centering
\small
\begin{tabularx}{\columnwidth}{@{}lX@{}}
\toprule
Metric & What it diagnoses \\
\midrule
MAE ($\downarrow$) & Primary euro-scale error metric. \\
RMSE ($\downarrow$), $R^2$ ($\uparrow$) & Sensitivity to extreme awards. \\
MedAE ($\downarrow$), 95AE ($\downarrow$) & Body and upper-tail error. \\
Pearson $r$ ($\uparrow$) & Linear association with award magnitude. \\
Spearman $\rho$ ($\uparrow$) & Rank ordering. \\
Zero-F1 ($\uparrow$) & Zero-award recognition. \\
\bottomrule
\end{tabularx}
\caption{Evaluation metrics.}
\label{tab:metrics}
\end{table}

\section{Results and Analysis}
\label{sec:results}

\subsection{Main Results}
\label{sec:main-results}

Table~\ref{tab:main-results} reports performance on the test pool and diagnostic subsets. Prompted decoder LM rows use the raw-text input unless otherwise specified; daggered rows are diagnostic settings with retrieved references or ReAct tool access.

\begin{table*}[!t]
\centering
\scriptsize
\setlength{\tabcolsep}{4pt}
\renewcommand{\arraystretch}{0.9}
\begin{tabular}{llrrrrrrrrrrr}
\toprule
Model & Variant & MAE$\downarrow$ & RMSE$\downarrow$ & $R^2$$\uparrow$ & MedAE$\downarrow$ & 95AE$\downarrow$ & Pearson $r$$\uparrow$ & Spearman $\rho$$\uparrow$ & Zero-F1$\uparrow$ & ID$\downarrow$ & OOD$\downarrow$ & Chal.$\downarrow$ \\
\midrule
\multicolumn{13}{l}{\textit{Constant baselines}} \\
Train median  & --- & 11{,}006 & 47{,}223 & -0.034 & 3{,}200 & 38{,}820 & 0.000 & 0.000 & 0.000 & 9{,}752 & 11{,}668 & 22{,}812 \\
Train mean    & --- & 14{,}940 & 46{,}468 & -0.001 & 11{,}010 & \textbf{29{,}010} & 0.000 & 0.000 & 0.000 & 14{,}465 & 15{,}190 & 25{,}150 \\
Constant zero & --- & 11{,}713 & 47{,}904 & -0.064 & 3{,}400 & 42{,}020 & 0.000 & 0.000 & 0.495 & 10{,}309 & 12{,}454 & 22{,}916 \\
\midrule
\multicolumn{13}{l}{\textit{Gradient-boosted trees}} \\
CatBoost & struct. feats. & \textbf{9{,}881} & 44{,}231 & 0.093 & \textbf{2{,}315} & 37{,}819 & 0.353 & \textbf{0.533} & 0.190 & \underline{8{,}143} & \underline{10{,}797} & 20{,}987 \\
XGBoost  & struct. feats. & 10{,}117 & 44{,}713 & 0.073 & 2{,}404 & 34{,}740 & 0.330 & \underline{0.523} & 0.123 & 8{,}180 & 11{,}138 & 21{,}527 \\
LightGBM & struct. feats. & 10{,}240 & 50{,}785 & -0.195 & \underline{2{,}395} & 33{,}020 & 0.304 & 0.518 & 0.104 & 9{,}312 & \textbf{10{,}729} & 21{,}085 \\
\midrule
\multicolumn{13}{l}{\textit{Retrieval baselines}} \\
$k$NN  & metadata        & 10{,}607 & 45{,}448 & 0.043 & 3{,}400 & 32{,}000 & 0.275 & 0.268 & 0.198 & 8{,}863 & 11{,}527 & 23{,}305 \\
$k$NN  & struct. feats. & 10{,}632 & 43{,}355 & 0.129 & 2{,}850 & 42{,}730 & 0.408 & 0.381 & \underline{0.517} & \textbf{8{,}140} & 11{,}945 & 24{,}665 \\
BM25   & raw text      & 13{,}390 & 49{,}744 & -0.147 & 4{,}500 & 43{,}540 & 0.254 & 0.335 & 0.505 & 11{,}494 & 14{,}389 & 22{,}980 \\
BGE-M3 & sparse          & 10{,}226 & 43{,}839 & 0.109 & 3{,}200 & 33{,}000 & 0.382 & 0.321 & \textbf{0.530} & 8{,}796 & 10{,}980 & \textbf{20{,}378} \\
BGE-M3 & dense           & 10{,}177 & 43{,}401 & 0.127 & 3{,}150 & 33{,}020 & 0.448 & 0.323 & 0.425 & 8{,}354 & 11{,}138 & 21{,}467 \\
\midrule
\multicolumn{13}{l}{\textit{Trained Encoder LMs}} \\
ModernBERT       & raw text  & 10{,}133 & \underline{42{,}232} & \underline{0.173} & 2{,}997 & 35{,}759 & \underline{0.476} & 0.329 & 0.209 & 8{,}438 & 11{,}026 & 21{,}544 \\
ModernBERT       & late fusion & \underline{10{,}074} & \textbf{41{,}579} & \textbf{0.199} & 2{,}999 & 36{,}391 & 0.475 & 0.312 & 0.067 & 8{,}404 & 10{,}955 & 21{,}024 \\
Legal-Longformer & raw text  & 10{,}642 & 46{,}619 & -0.007 & 2{,}695 & 38{,}461 & 0.184 & 0.322 & 0.017 & 9{,}200 & 11{,}402 & 22{,}116 \\
Legal-Longformer & late fusion & 10{,}244 & 42{,}439 & 0.165 & 2{,}760 & 36{,}021 & 0.426 & 0.487 & 0.226 & 8{,}695 & 11{,}060 & 21{,}013 \\
\midrule
\multicolumn{13}{l}{\textit{Prompted Decoder LMs (zero-shot)}} \\
Qwen3.5-9B   & raw text  & 16{,}598 & 110{,}079 & -4.616 & 6{,}750 & 35{,}000 & 0.011 & 0.190 & 0.196 & 17{,}113 & 16{,}327 & 24{,}212 \\
Qwen3.5-27B  & raw text & 22{,}235 & 79{,}735 & -1.947 & 7{,}500 & 125{,}000 & 0.442 & 0.211 & 0.094 & 17{,}584 & 24{,}687 & 52{,}066 \\
Qwen3.5-Plus & raw text & 16{,}400 & 68{,}894 & -1.199 & 5{,}000 & 51{,}575 & 0.451 & 0.290 & 0.115 & 12{,}781 & 18{,}306 & 37{,}186 \\
GPT-OSS-20B  & raw text & 15{,}678 & 67{,}390 & -1.105 & 6{,}000 & 52{,}000 & 0.076 & 0.072 & 0.425 & 13{,}363 & 16{,}898 & 23{,}973 \\
GPT-5.4      & raw text & 25{,}438 & 85{,}330 & -2.375 & 4{,}800 & 104{,}000 & 0.135 & 0.249 & 0.006 & 17{,}140 & 29{,}812 & 47{,}613 \\
\midrule
\multicolumn{13}{l}{\textit{Prompted Decoder LMs (zero-shot CoT)}} \\
Qwen3.5-9B   & raw text & 10{,}782 & 44{,}551 & 0.080 & 4{,}500 & \underline{31{,}005} & 0.317 & 0.183 & 0.255 & 9{,}777 & 11{,}312 & \underline{20{,}926} \\
Qwen3.5-27B  & raw text & 12{,}905 & 43{,}497 & 0.123 & 5{,}300 & 44{,}945 & 0.440 & 0.256 & 0.035 & 9{,}848 & 14{,}516 & 26{,}729 \\
Qwen3.5-Plus & raw text & 19{,}080 & 75{,}337 & -1.631 & 6{,}500 & 67{,}600 & \textbf{0.528} & 0.291 & 0.029 & 13{,}220 & 22{,}170 & 42{,}135 \\
GPT-OSS-20B  & raw text & 16{,}748 & 59{,}098 & -0.618 & 6{,}500 & 56{,}100 & 0.153 & 0.178 & 0.069 & 13{,}468 & 18{,}478 & 31{,}543 \\
GPT-5.4      & raw text & 16{,}266 & 48{,}458 & -0.088 & 5{,}000 & 72{,}000 & 0.452 & 0.271 & 0.019 & 10{,}754 & 19{,}172 & 39{,}884 \\
\midrule
\multicolumn{13}{l}{\textit{Prompted Decoder LMs (few-shot CoT)$^\dagger$}} \\
Qwen3.5-9B   & 5 retrieved$^\dagger$ & 14{,}867 & 58{,}935 & -0.610 & 5{,}000 & 50{,}700 & 0.236 & 0.201 & 0.095 & 11{,}127 & 16{,}836 & 33{,}169 \\
Qwen3.5-Plus & 5 retrieved$^\dagger$ & 24{,}141 & 79{,}094 & -1.900 & 6{,}000 & 105{,}000 & 0.277 & 0.235 & 0.076 & 13{,}518 & 29{,}741 & 60{,}814 \\
GPT-OSS-20B  & 5 retrieved$^\dagger$ & 11{,}674 & 46{,}410 & 0.002 & 5{,}000 & 36{,}725 & 0.171 & 0.155 & 0.065 & 10{,}007 & 12{,}553 & 24{,}688 \\
GPT-5.4      & 5 retrieved$^\dagger$ & 22{,}051 & 143{,}132 & -8.495 & 5{,}500 & 84{,}000 & 0.126 & 0.253 & 0.035 & 13{,}342 & 26{,}642 & 48{,}864 \\
\midrule
\multicolumn{13}{l}{\textit{Knowledge-augmented agents (ReAct)$^\dagger$}} \\
Qwen3.5-Plus &  ReAct$^\dagger$ & 18{,}220 & 55{,}386 & -0.420 & 5{,}500 & 65{,}000 & 0.149 & 0.204 & 0.002 & 11{,}156 & 21{,}946 & 47{,}497 \\
MiniMax-M2.7 &  ReAct$^\dagger$ & 16{,}571 & 52{,}914 & -0.298 & 5{,}000 & 59{,}840 & 0.172 & 0.199 & 0.029 & 10{,}911 & 19{,}555 & 40{,}758 \\
\bottomrule
\end{tabular}
\caption{Main results on the ECtHR-NPD test pool. Rows follow the input settings unless marked with $^\dagger$ for diagnostic settings using additional signals, such as retrieved few-shot references or ReAct tool access. For each metric, \textbf{bold} marks the best value across the full table and underlining marks the second-best value.}
\label{tab:main-results}
\end{table*}

\paragraph{Aggregate performance on MAE.}
Trained baselines (gradient-boosted trees and encoder LMs) achieve lower MAE than prompted decoder LM variants and ReAct agents. CatBoost obtains the lowest aggregate MAE (\EUR{9.9}k), which is 10.2\% lower than the training-set median. Prompt-based decoder LMs perform less reliably. The best prompted LM result is Qwen3.5-9B with static CoT (\EUR{10.8}k MAE), while vanilla zero-shot, retrieved few-shot CoT, and the ReAct setting remain worse than trained baselines. Larger LMs or complex agentic configurations do not consistently improve MAE. This is consistent with prior findings that LMs treat numbers as token sequences and struggle with magnitude-sensitive prediction~ \citep{spithourakis2018numeracy,li2025exposing}.

\paragraph{Secondary metrics.}
 Most systems have low or negative $R^2$, showing that they explain little variance on the original euro scale and are strongly affected by rare high-award cases. Rank metrics are more favourable for some supervised baselines like CatBoost and ModernBERT late fusion. This suggests that some models recover ordering better than predict magnitudes. Retrieval-based methods achieve relatively high Zero-F1 scores because they can recognise some zero-award cases, but this does not imply good positive-award prediction. 

\paragraph{Bootstrap tests for performance gaps.}
We use full-size paired bootstrap tests to distinguish stable leaderboard gaps from close numerical differences (Appendix~\ref{app:bootstrap-significance}). The tests show that CatBoost is significantly better than the train-median and the strongest prompted LM, but not significantly better than BGE-M3 dense or ModernBERT late fusion. Prompting gains remain internal to the decoder-LM family, rather than reliable benchmark-level gains over the train median.

\subsection{Input Representation Ablations}
\label{sec:input-calibration-ablations}

\paragraph{Structured input may help.}
Across model families, structured representations appear useful when
they compress long judgments into legally relevant signals. Encoder late fusion gives small but consistent improvements over text-only fine-tuning. Similarly, serialising the structured features extracted from the Facts section improves several models in zero-shot and CoT settings. This suggests that part of the difficulty for prompted LMs is the long-context reading burden, as the serialisation compresses the average case input from thousands of tokens to hundreds. At the 95th percentile, input length falls from 5{,}337 words for the raw-text representation to 237 words for the structured representation (Appendix Table~\ref{tab:input-length-comparison}). However, these gains are modest and unstable. The input compression alone does not solve case-specific reasoning or reliable calibration of award magnitude.

\paragraph{Adding more permitted case information has diminishing returns.}
The tree and prompting ablations show that adding more permitted signals to the model input does not yield monotonic gains. In the tree ablations
(Table~\ref{tab:tree-feature-ablation}), moving from a compact metadata-based input to increasingly richer case representations does not consistently improve results. Prompted LMs show a similar pattern, where CoT and retrieved few-shot examples help some models but hurt others. Thus, the results indicate that mapping facts to realised euro amounts remains a bottleneck for the evaluated systems.  These results show that incorporating additional case information does not consistently reduce monetary prediction error.

\begin{table}[!t]
\centering
\scriptsize
\setlength{\tabcolsep}{3pt}
\renewcommand{\arraystretch}{1.0}
\resizebox{\columnwidth}{!}{%
\begin{tabular}{llrrr}
\toprule
Model & Prompting & Raw & Struct. & $\Delta$ \\
\midrule
Qwen3.5-9B & zero-shot & 16{,}598 & 13{,}218 & -3{,}380 \\
Qwen3.5-27B & zero-shot & 22{,}235 & 18{,}118 & -4{,}117 \\
Qwen3.5-9B & CoT & 10{,}782 & 11{,}815 & +1{,}033 \\
Qwen3.5-Plus & CoT & 19{,}080 & 16{,}602 & -2{,}478 \\
GPT-OSS-20B & CoT & 16{,}748 & 11{,}707 & -5{,}041 \\
GPT-5.4 & CoT & 16{,}266 & 14{,}384 & -1{,}882 \\
Qwen3.5-9B & few-shot CoT$^\dagger$ & 14{,}867 & 14{,}409 & -458 \\
GPT-OSS-20B & few-shot CoT$^\dagger$ & 11{,}674 & 12{,}074 & +400 \\
\bottomrule
\end{tabular}
}
\caption{Raw versus serialised inputs for prompted decoder LMs. $\Delta$ is
structured MAE minus raw-text MAE; negative values are better.
$^\dagger$ denotes diagnostic few-shot settings.}
\label{tab:serialization}
\end{table}

\paragraph{Award-related information is easy for decoder LMs to exploit.}
Our diagnostic runs show why award-related signals must be separated from model input. Claimed amounts can act as direct anchors or practical ceilings under Rule~60 and the Court's just-satisfaction procedure
\citep{echr2025rules, echr2022justsatisfaction}, and
Article~41 reasoning often reveals whether a case has a zero award. Adding this information roughly halves MAE for both agents and raises Zero-F1 to approximately 0.99 (Table~\ref{tab:agent-input-diagnostics}). This shows that decoder LMs and agents perform substantially better when given explicit award-related information than when they must infer the award from case facts.

\begin{table}[!t]
\centering
\small
\setlength{\tabcolsep}{4pt}
\renewcommand{\arraystretch}{1.05}
\begin{tabular}{llrr}
\toprule
Model & Setting & MAE & Zero-F1 \\
\midrule
Qwen3.5-Plus
& base
& 18{,}220
& 0.002 \\
Qwen3.5-Plus
& expanded
& 8{,}810 ($\downarrow$51.6\%)
& 0.997  \\
MiniMax-M2.7
& base 
& 16{,}571
& 0.029 \\
MiniMax-M2.7
& expanded
& 8{,}819 ($\downarrow$46.8\%)
& 0.988  \\
\bottomrule
\end{tabular}
\caption{ReAct results using the benchmark inputs (Base) and with applicants’ claims and Article~41 reasoning added (Expanded). The latter is a diagnostic setting and is not included in the main model comparison.}
\label{tab:agent-input-diagnostics}
\end{table}

\subsection{Distributional Shift and Challenging Cases}
\label{sec:temporal-legal-shift}

\paragraph{Validation-to-test shift.}
Trained and fine-tuned models are selected using the validation split, but the later test period
differs structurally from validation (Table~\ref{tab:val-test-structural-shift}). These compositional changes may partly explain why validation-fit models do not transfer cleanly to the test period.

\paragraph{Results by Diagnostic view.}

MAE for the ID subset is lower than that for the OOD subset for both trained models and prompted LMs (Table~\ref{tab:main-results}). The prompted LMs receive no task-specific fine-tuning. This suggests that the ID set is not only distributionally closer to the training period but may also be simpler on average. All model families also degrade sharply on the challenging view. Across representative systems, challenging-set MAE is roughly twice full-test MAE. The Challenging subset also differs in respondent state, zero-award prevalence, applicant count, violation count, and violated Articles, so higher error reflects multiple sources of difficulty.

\subsection{Error Analysis}
\label{sec:error-structure}
Aggregate MAE hides systematic differences across the award distribution and across legally meaningful case groups. We therefore decompose errors by award range, respondent state, and violated Article.

\begin{table}[!t]
\centering
\small
\setlength{\tabcolsep}{5pt}
\renewcommand{\arraystretch}{1.05}

\resizebox{\columnwidth}{!}{%
\begin{tabular}{lrrr}
\toprule
Dimension & Val. & Test & $\Delta$ \\
\midrule
Zero-award cases & 25.1\% & 32.9\% & +7.8  \\
Committee judgments & 65.2\% & 75.1\% & +9.9  \\
Multiple-applicant cases & 36.1\% & 45.8\% & +9.7  \\
High applicant-count cases ($\geq$6) & 10.9\% & 19.3\% & +8.4  \\
Complex violation cases ($\geq$3) & 19.0\% & 26.7\% & +7.7  \\
\bottomrule
\end{tabular}
}
\caption{Validation-to-test structural shift (\%; $\Delta$ is the test percentage minus the validation percentage, in percentage points).}
\label{tab:val-test-structural-shift}
\end{table}

\paragraph{Award-range errors.}
Table~\ref{tab:award-range-errors} separates errors by target-award
range. For zero-award cases, prompted LMs and ReAct diagnostics
often predict positive awards. This is expected under the input setting, which hides the Court's Article~41 reasoning and procedural
signals that often explain zero awards. For high-award cases, all
representative systems show large magnitude errors. Although awards above \EUR{50}k account for only 4.1\% of the test pool (118 of 2{,}897 cases),
excluding them substantially lowers MAE for
every representative system in Table~\ref{tab:award-range-errors}. Frequent zero awards and the small number of very large awards therefore affect aggregate performance in different ways.

\paragraph{Results by respondent state and violated Article.}
Errors and award distributions also vary by respondent state and violated Article (Appendix~\ref{app:state-article-side-views}, Tables~\ref{tab:state-docket-side-view} and \ref{tab:article-side-view}). These patterns do not mean that cases from a particular respondent state or involving a particular Convention article receive higher or lower awards. The observed differences may also reflect the composition of cases within each group, including the prevalence of zero awards, multi-applicant and bundled-award structures, combinations of violations, and the nature of the harm involved.

For example, Russian cases often contain many multi-applicant judgments and zero-award cases, both of which can affect the observed award distribution. Article~2 concerns the right to life, and cases under this Article often involve death or serious harm, which may partly explain the larger awards observed in this group. Article patterns can also be affected by co-occurring violations. For example, Article~13 is applied in connection with claims under other Convention rights, so award patterns in this article may also reflect the other rights and harms involved in those cases. State-level and Article-level patterns therefore need to be interpreted together with other characteristics of the cases.

\section{Discussion}
\label{sec:discussion}

\subsection{Model Behaviour and Prior Knowledge}
\label{sec:anchors-calibration}

The ablations in Section~\ref{sec:input-calibration-ablations} show that
serialised inputs, CoT prompting, retrieved examples, and ReAct knowledge access can all change prompted LMs' behaviour. These interventions provide a reasoning path, prior cases with their awards, or a legal knowledge base. However, they do not reliably solve the zero/positive recognition or the calibration of award magnitude. This explains why prompting gains remain mostly internal to the decoder-LM family and do not become stable benchmark-level gains over the train-median baseline.

\paragraph{Multi-applicant aggregation.}
Legal reasoning traces often look plausible while remaining numerically
miscalibrated. For cases in the Challenging set, the target is usually a holistic case-level award rather than a transparent
sum of applicant-level awards or violation-level harms. CoT and ReAct
traces encourage models to discuss severity, applicant count, violation
subtype, and comparable case factors, but they do not recover the
Court's implicit award scale. In some multi-applicant cases, the model
over-scales the prediction as if NPD awards were linear per-applicant
sums. This can produce very large errors, as shown in Table~\ref{tab:multi-applicant-cot-failures}.

\begin{table}[!htbp]
\centering
\small
\setlength{\tabcolsep}{4pt}
\renewcommand{\arraystretch}{1.08}
\begin{tabular}{lrrr}
\toprule
Case ID & Applicants & \shortstack{Ground\\truth} & \shortstack{Model\\prediction} \\
\midrule
001-219675 & 90  & 12{,}500 & 450{,}000 \\
001-241738 & 195 & 7{,}500  & 1{,}500{,}000 \\
\bottomrule
\end{tabular}
\caption{Two Qwen3.5-Plus examples illustrating that linear scaling with applicant count leads to large prediction errors. Amounts are in EUR.}
\label{tab:multi-applicant-cot-failures}
\end{table}

\subsection{Legal and Numerical Challenges in Monetary Prediction}
\label{sec:magnitude-aware-models}

Our results identify several recurring challenges for this benchmark, including distinguishing zero from positive awards, combining information across applicants and violations, and estimating the magnitude of rare high awards. These require both identifying legally relevant information and converting it into an appropriate case-level monetary amount.

The challenges mentioned above are not only failures of legal reasoning. They also reflect a numerical calibration problem. Standard LM training objectives treat numbers as token sequences and do not directly optimise monetary distance. Errors of \EUR{5,000} and \EUR{500,000} are not naturally penalised in proportion to their legal and economic difference. Future research
on ECtHR-NPD will likely require approaches designed for magnitude
prediction, such as hurdle models~\citep{cragg1971hurdle,liu2020deephurdle},
quantile heads~\citep{koenker1978,wang2025quantilellm}, distributional
regression~\citep{rigby2005additive,kneib2023distribution}, or
number-aware losses that penalise predictions by numerical distance
\citep{zausinger2025regress}. A two-stage hurdle model first identifies
whether the award is zero and then predicts the amount for positive awards. It would be difficult to use under the current model input setting because reasons for zero awards in Article~41 are withheld. We retain direct case-level regression in this paper so heterogeneous systems are compared on the same output. 

\begin{table}[!t]
\centering
\small
\setlength{\tabcolsep}{3pt}
\renewcommand{\arraystretch}{1.03}

\begin{tabular}{@{}lrrrr@{}}
\toprule
Model & Zero & $>$0--10k & 10k--50k & $>$50k \\
\midrule
CatBoost & 3{,}089 & 2{,}401 & 15{,}486 & 125{,}816 \\
BGE-M3 sparse & 2{,}858 & 2{,}567 & 16{,}010 & 132{,}087 \\
ModernBERT LF & 3{,}367 & 2{,}484 & 15{,}722 & 126{,}386 \\
Qwen3.5-9B CoT & 6{,}597 & 3{,}713 & 13{,}404 & 113{,}482 \\
GPT-OSS few-shot$^\dagger$ & 7{,}926 & 3{,}077 & 14{,}791 & 126{,}038 \\
MiniMax ReAct$^\dagger$ & 21{,}977 & 6{,}345 & 13{,}121 & 102{,}854 \\

\bottomrule
\end{tabular}
\caption{MAE by target-award range for representative systems. Values are
in EUR. The buckets separate boundary errors at zero from magnitude errors in the high-award tail.}
\label{tab:award-range-errors}
\end{table}

\subsection{Monetary Remedies Require Calibration-Aware Evaluation}
\label{sec:legal-nlp-regression-gap}

ECtHR-NPD tests whether models can predict realised Court awards from permitted case inputs. Existing legal NLP benchmarks do not
capture the main failure modes exposed here: zero-award recognition,
high-tail magnitude error, and degradation under violation/applicant aggregation and distribution shift. Calibration-aware evaluation for monetary remedies requires reporting award-range errors, zero-award recognition, diagnostic views, and legally meaningful side views rather than relying only on aggregate accuracy or aggregate MAE.

\section{Conclusion}
\label{sec:conclusion}

We introduced \textbf{ECtHR-NPD}, a benchmark for predicting
Article~41 non-pecuniary damage awards at the European Court of Human
Rights. ECtHR-NPD provides 14{,}575 validated targets,
chronological splits, diagnostic test views, and a clear separation of target construction from model input. Our results show that ECtHR-NPD remains challenging for evaluated open-weight and proprietary LMs. More broadly, monetary remedy prediction exposes failures that are not measured by existing legal NLP benchmarks.

ECtHR-NPD extends legal NLP evaluation to continuous monetary remedies and
provides a testbed for studying the challenges of monetary prediction. Future
work can explore improved numerical modelling, two-stage and distributional
approaches, prompt optimisation, uncertainty and calibration, legally
meaningful evaluation ranges, and how models select evidence and explain
monetary predictions.

\section*{Limitations}

\paragraph{Project Scope.} 
ECtHR-NPD focuses on Article~41 non-pecuniary damage awards in English-language ECtHR judgments and uses case-level targets. It does not cover non-English judgments, pecuniary damages, costs and expenses, per-applicant award prediction, or other international and domestic courts. Findings should therefore not be generalised beyond this institutional setting without further validation.

\paragraph{Target construction.}
The targets are produced by a multi-source extraction pipeline with deterministic semantic contracts (Section~\ref{sec:target-construction} and Appendix~\ref{app:target-construction}). Although we perform reconciliation and validation checks, residual label noise may remain, especially in older judgments and multi-applicant cases with bundled awards.

\paragraph{Equitable awards and point-error metrics.}
Article~41 awards are determined on an equitable basis: on the same facts, several awards within a plausible range may be legally defensible. Point-error metrics treat any deviation from the realised award as wrong, including deviations that fall inside this defensible range. ECtHR-NPD therefore measures fit to the Court's realised practice rather than legal correctness, and future evaluation should consider band-level or range-based protocols alongside absolute error.

\paragraph{Evaluation under heavy-tailed targets.}
Evaluation is difficult because the target is zero-inflated, heavy-tailed, and time-sensitive. MAE is interpretable but may understate upper-tail failures, while RMSE and R$^2$ are dominated by rare extreme cases and rank metrics do not measure monetary calibration. In addition, euro-denominated awards reflect inflation, historical currency conversion, shifts in Court practice, and institution-specific procedures. The benchmark measures prediction of the Court's historical practice, not normatively just outcomes.

\paragraph{Computation cost and LM reproducibility.}
Predictions from prompted decoder LMs and ReAct agents are not perfectly deterministic. Although we use matched prompting settings where possible and set decoding temperature to zero, LM execution may still introduce run-to-run variation. Our budget and compute constraints prevent us from repeating all prompted decoder-LM and ReAct runs to obtain averaged scores. We therefore treat prompted decoder-LM results as single-run estimates under a fixed protocol rather than fully variance-characterised measurements.

\section*{Ethics Statement}

\paragraph{Intended use.}
This work is intended as a benchmark for legal NLP and empirical legal analysis, not as a system for legal advice, settlement anchoring, or automated judicial decision-making. Model predictions reflect historical ECtHR practice and may reproduce institutional patterns or biases in that practice. Compensation in international human rights law also has a recognitional function, shaping which injuries, victims, and forms of suffering are legally acknowledged~\citep{gonzalezsalzberg2022queering}. Models trained on historical awards may therefore reproduce existing patterns of recognition and exclusion. They should not be used to advise applicants, inform court decisions, or substitute for expert legal review.

\paragraph{Data source and release.}
ECtHR-NPD is constructed from publicly available HUDOC judgments. We do not redistribute raw judgment text. The released package contains validated targets, leakage-audited structured features, split and diagnostic-view tags, and public HUDOC case identifiers, allowing researchers to reproduce document retrieval from official sources under ECtHR terms of use. 

\paragraph{Sensitive case contexts and representativeness.}
The released benchmark excludes applicant-level identifying information and avoids redistributing raw judgment text. However, the underlying ECtHR cases still involve sensitive legal and personal contexts, including detention, displacement, torture or ill-treatment, bereavement, and interference with private or family life. The dataset should therefore be handled with care and used only for research purposes, not for profiling applicants or supporting real-world compensation decisions. The observed docket is not a balanced sample of legal questions or affected populations. In particular, the post-2021 test window contains a disproportionate share of Russia- and Ukraine-related cases. We preserve this composition for benchmark fidelity, but interpret results in light of this distribution shift.

\paragraph{Use of AI assistants.}

AI assistants were used to support code debugging, draft editing, and experiment-log summarisation. All substantive claims, interpretations, experimental results, and final text were reviewed and verified by the authors.

\section*{Acknowledgments}

We would like to thank Mingzi Cao and Vynska Amalia Permadi for their helpful internal reviews. We acknowledge IT Services at the University of Sheffield for the provision of services for High Performance Computing. YP is supported by the UKRI AI Centre for Doctoral Training in Speech and Language Technologies (SLT) and their Applications, funded by UK Research and Innovation [grant number EP/S023062/1]. ZY is partly supported by The Alan Turing Institute through the Development of an AI Data Engineer project. NA is partly supported by the EPSRC [grant number EP/Y009800/1] through funding from Responsible AI UK (KP0016) as a Keystone project. 

\bibliography{references}

\clearpage
\appendix
\label{sec:appendix}

\section{Legal Background for the ECtHR-NPD Benchmark}
\label{app:legal-background}

This appendix provides the legal background necessary for readers without
prior familiarity with the European Convention on Human Rights (ECHR) to
follow the prediction target, leakage control, and modelling choices in
ECtHR-NPD. It is intended as a bridge for AI and NLP readers, not as a doctrinal exposition. The appendix introduces the
Convention system and Article~41 (\S\ref{app:convention}), the structure
of ECtHR judgments and the resulting leakage boundary
(\S\ref{app:structure}), the distinction between pecuniary and
non-pecuniary damage and the zero-award taxonomy (\S\ref{app:taxonomy}),
the article-level and aggregation issues that motivate diagnostic
reporting (\S\S\ref{app:articles}--\ref{app:multi}), and the Court's equitable basis for determining the award amount (\S\ref{app:equity}).

\subsection{The Convention system and Article 41}
\label{app:convention}

The European Court of Human Rights (ECtHR or ``Court'') in Strasbourg hears applications alleging violations of the European Convention on Human Rights~\citep{echr1950convention}. Individuals who consider themselves victims of a Convention violation may apply directly to the Court once domestic remedies have been exhausted (Article~35). Where the Court finds a violation, Article 41 of the Convention allows it to ``afford just satisfaction to the injured party'' against the respondent state. Just satisfaction is conventionally divided into three heads: pecuniary damage (PD), non-pecuniary damage (NPD), and costs and expenses~\citep{echr2022justsatisfaction}. Older judgments may refer to the predecessor just-satisfaction provision Article~50. We apply the same target-construction procedure to historical Article~50 cases only when the non-pecuniary award can be reliably identified; unresolved historical Article~50 or pre-euro currency cases are excluded.

The benchmark targets NPD because it captures intangible harm described in the judgment facts. Unlike pecuniary damage and costs, NPD is not primarily evidenced through receipts, invoices, or proof of financial loss. The Court instead assesses NPD on an equitable basis, drawing on the facts and consequences of the violation. NPD is therefore a continuous monetary target whose amount is not fixed by a public statutory formula or schedule.

\subsection{Structure of an ECtHR judgment}
\label{app:structure}

ECtHR judgments have a recurrent structure that lets us separate case information from award-related material used to establish the target. Not every judgment follows this layout exactly~\citep{echr2022hudocmanual}, so our extraction pipeline accommodates other variants. Figure~\ref{fig:judgment-structure} maps the common structure of an ECtHR judgment to the model input policy.

\begin{figure}[t]
\centering
\small
\setlength{\fboxsep}{2pt}
\begin{tikzpicture}[
  box/.style={
    draw,
    rounded corners,
    align=left,
    text width=0.92\columnwidth,
    inner sep=3pt,
    minimum height=0.72cm
  }
]

\node[box, fill=gray!8] (procedure) at (0,0) {
\textbf{Procedure.}
Application history, parties, Chamber composition, hearings, and third-party interventions.
};

\node[box, fill=blue!8, below=0.12cm of procedure] (facts) {
\textbf{The Facts.}
Circumstances of the case, relevant domestic law and practice, and relevant international or comparative material.
\textit{Main textual input in the raw-text representation.}
};

\node[box, fill=orange!10, below=0.12cm of facts] (law) {
\textbf{The Law.}
Admissibility, merits reasoning, and conclusions on alleged Convention violations.
\textit{Raw text excluded; the finding of violated articles is provided to the model.}
};

\node[box, fill=red!8, below=0.12cm of law] (a41) {
\textbf{Application of Article~41.}
Applicant claims, government observations, compensation reasoning, and award discussion.
\textit{Label construction and validation only.}
};

\node[box, fill=red!8, below=0.12cm of a41] (operative) {
\textbf{Operative Provisions.}
Formal holdings, often introduced by ``For these reasons, the Court'', including payment orders.
\textit{Label construction and validation only.}
};

\end{tikzpicture}
\caption{Common structure of an ECtHR judgment and the model input policy. The raw-text representation uses the \textsc{Facts} section, and structured models use structured case information. Article~41 material, operative payment provisions, applicant claims, and award tables are used only to construct and validate targets.}
\label{fig:judgment-structure}
\end{figure}

The model-input policy follows this judgment structure. The raw-text representation uses the \textsc{Facts} section, and structured models use structured case information. Article~41 material, operative payment provisions, applicant claims, and award tables are used only to construct and validate targets. This separation prevents models from extracting the award amount directly from award-related sections and allows them to predict NPD awards based on
case information and controlled merits findings.

Judgments may be delivered by Committees, Chambers, or the Grand
Chamber. These formations differ in procedural role and case composition,
so we record court formation as a structural feature. From a legal perspective, Grand Chamber cases often concern questions of particular importance or complexity, and we therefore include them in the Challenging diagnostic view. Under Article~28, three-judge Committees may decide repetitive cases where the underlying question is already covered by well-established case law. A single judgment may contain multiple \textbf{The Law} sub-sections (one per alleged Article) and multiple Article~41 sub-determinations.

\subsection{Pecuniary versus non-pecuniary damage, and the zero-award taxonomy}
\label{app:taxonomy}

The distinction between pecuniary and non-pecuniary damage is substantive rather than terminological: the two heads rely on different evidentiary bases and forms of legal assessment.

\paragraph{Pecuniary damage} compensates quantifiable material loss, such as lost earnings, medical costs, or property damage. The Court requires a causal link between the violation and the loss claimed, and pecuniary claims are normally substantiated by documentary evidence \citep[\S\S10--13]{echr2022justsatisfaction}. We exclude pecuniary damage from the target because it is driven by financial evidence and less directly tied to the intangible harm described in the facts.

\paragraph{Non-pecuniary damage} compensates suffering, distress, anxiety, frustration, feelings of injustice, loss of reputation, and analogous intangible harms. Unlike pecuniary damage, NPD is less directly tied to documentary proof of financial loss~\citep[\S\S14--15]{echr2022justsatisfaction}. The Court may award an amount on an equitable basis (\texttt{equitable\_basis}), declare that the finding of a violation constitutes sufficient just satisfaction (\texttt{finding\_sufficient}), or treat compensation already awarded at the domestic level as sufficient financial redress (\texttt{domestic\_award\_cover}).

Many ECtHR cases receive zero NPD awards for reasons that are unrelated to the severity of the violation, as listed below:

\begin{enumerate}[leftmargin=2em, topsep=2pt, itemsep=0pt]
    \item \texttt{finding\_sufficient}. The Court states that the
    finding of a violation constitutes sufficient just satisfaction.
    This is common where the Court recognises the violation but does not
    consider a separate monetary NPD award necessary.

    \item \texttt{no\_claim}. The applicant did not submit a
    non-pecuniary damage claim. Under Rule~60 of the Rules of Court,
    applicants must submit itemised just-satisfaction claims in due form
    and within the applicable time limit~\citep{echr2025rules}.

    \item \texttt{unsubstantiated}. The applicant claimed
    non-pecuniary damage, but the Court rejected the claim because it was
    not sufficiently substantiated.

    \item \texttt{rule\_60\_non\_compliance}. The just-satisfaction
    claim failed to comply with Rule~60 requirements, for example, because
    it was submitted late, was not itemised, or was otherwise procedurally defective.

    \item \texttt{domestic\_award\_covers}. Domestic compensation or
    other domestic redress was treated as sufficient financial redress for the relevant non-pecuniary harm, so the Court did not make a further NPD award.

    \item \texttt{applicant\_deceased\_no\_heir}. The applicant died
    and no heir, relative, or continuing applicant pursued the
    just-satisfaction claim.
\end{enumerate}

These categories show that a zero award can reflect procedural or remedial considerations rather than a less serious underlying violation. Many of those considerations are recorded in the Court's Article~41 reasoning, which is deliberately withheld from model input to prevent target leakage. A zero award is therefore neither a proxy for lower severity nor simply the lower end of a continuous monetary target. The task thus combines two questions: whether a monetary award is made and, if so, how large it is. Distinguishing zero from positive awards is therefore a particularly demanding diagnostic for \textsc{ECtHR-NPD}.

\subsection{Article-level harm structures relevant to NPD prediction}
\label{app:articles}

NPD does not follow one universal severity scale across Convention provisions. The Court’s equitable assessment instead turns on different factual axes for different rights. Table~\ref{tab:article_conventions} summarises the protected interest and NPD-relevant axes for the main Convention provisions appearing in the benchmark and diagnostic analysis. These axes are used to motivate diagnostics and interpretation; they are neither gold labels nor prescribed rules for prediction.

\begin{table*}[t]
  \centering
  \small
  \setlength{\tabcolsep}{4pt}
  \renewcommand{\arraystretch}{1.15}
  \begin{tabular}{p{0.14\linewidth} p{0.31\linewidth} p{0.49\linewidth}}
    \toprule
    \textbf{Provision} & \textbf{Protected interest / typical harm} & \textbf{NPD-relevant axes} \\
    \midrule
    Article~2 &
    Life; death, lethal force, failure to protect life, ineffective investigation &
    Loss of life, threat to life, relationship between victim and applicant, and whether the violation is substantive, procedural, or both. \\

    Article~3 &
    Torture, inhuman or degrading treatment, detention conditions, removal-risk cases &
    Severity and duration of ill-treatment, physical and psychological harm, vulnerability, detention context, and distinction between substantive and procedural violations. \\

    Article~5 &
    Liberty and security; unlawful detention, excessive detention, lack of judicial review &
    Duration and type of detention, procedural safeguards, availability of review, and consequences for the applicant. \\

    Article~6 &
    Fair trial; access to court, length of proceedings, enforcement delays &
    Type of procedural unfairness, length and importance of proceedings, criminal or civil context, enforcement delay, and practical consequences for the applicant. \\

    Article~8 &
    Private life, family life, home, correspondence, reputation, privacy, and data protection &
    Nature of the interference, family separation, home interference, reputational or privacy harm, data-related impact, and applicant vulnerability. \\

    Article~10 &
    Freedom of expression; sanctions, chilling effects, journalist or political speech &
    Nature of the expression, public-interest context, severity of sanction, chilling effect, professional consequences, and status of the speaker. \\

    Article~11 &
    Freedom of assembly and association; protests, associations, political parties, unions &
    Nature of the assembly or association, political context, severity of interference, organisational impact, and whether applicants are individuals or organisations. \\

    Article~13 &
    Effective remedy &
    Relationship to the underlying violation, availability and effectiveness of domestic remedies, and whether the lack of remedy adds independent procedural harm. \\

    Article~14 &
    Non-discrimination, usually read with another Convention right &
    Protected ground, unequal-treatment context, dignitary harm, relationship to the underlying right, and practical consequences for the applicant. \\

    Protocol~1 Article~1 &
    Property and possessions; expropriation, non-enforcement, control of use &
    Nature and duration of property interference, uncertainty, enforcement delay, and separation between pecuniary loss and non-pecuniary harm. \\

    Protocol~1 Article~3 &
    Free elections &
    Nature of electoral interference, voting or candidacy rights, democratic participation, political context, and practical consequences. \\

    Article~9 &
    Thought, conscience, and religion &
    Nature of religious or conscience-based interference, institutional context, personal impact, and severity of restriction. \\

    Article~34 &
    Individual petition before the ECtHR &
    Obstruction of access to the Court, intimidation or pressure on applicants or representatives, procedural consequences, and relationship to other violations. \\

    Protocol~7 Article~2 &
    Right of appeal in criminal matters &
    Loss of appellate review, criminal-procedure context, consequences of conviction or sentence, and availability of alternative safeguards. \\

    Protocol~7 Article~4 &
    Ne bis in idem &
    Repeated prosecution or punishment, procedural burden, consequences for the applicant, and relationship to criminal penalties. \\
    \bottomrule
  \end{tabular}
  \caption{Article-level harm structures relevant to non-pecuniary damage prediction. The table summarises protected interests and the NPD-relevant axes for each Convention provision relevant to this paper. The Convention provisions and protected interests follow the Convention text~\citep{echr1950convention}; the NPD-relevant axes are distilled, for each provision, from the corresponding Case-Law Guide in the Court's Case-Law Guides series~\citep{echr2025caselawguides}, complemented by the Practice Direction on Just Satisfaction Claims~\citeyearpar{echr2022justsatisfaction}.}
  \label{tab:article_conventions}
\end{table*}

Several Convention rights distinguish substantive from procedural limbs: a particular Article can be violated through the underlying conduct (the killing, the ill-treatment) and through the failure to investigate that conduct. Where both limbs are present, the judgment does not necessarily indicate how much of the NPD amount corresponds to each limb. We therefore record both limbs explicitly, while keeping the supervised target at the case-level NPD award amount. 

In addition to the violated Convention articles, the Court also takes account of the applicant's position, the overall context of the breach, and local economic circumstances in the respondent State~\citep[\S12, \S14]{echr2022justsatisfaction}. Awards in similar cases may therefore vary across respondent States and over time.   We therefore include state-year GDP (as a proxy for respondent-state fiscal capacity) and GDP per capita (as a proxy for the local economic value of awards) as model inputs. For years through 2024, we obtain these covariates from the World Bank's World Development Indicators~\citep{worldbank2026wdi}; we use
IMF World Economic Outlook forecasts to extend the series for 2025--2026~\citep{imf2026weo}. Structured-input models use $\log(1+x)$-transformed versions.

\subsection{Multi-applicant, multi-violation, and joined cases}
\label{app:multi}

The prediction target in this benchmark is the total NPD amount awarded in each judgment. Many cases in ECtHR may address multiple applicants, multiple violation findings, or multiple joined applications while stating one global NPD amount. Such an amount cannot be
allocated reliably to a particular applicant, violation finding, or factual episode.

\paragraph{Multi-applicant cases.} A single judgment may contain multiple applicants, with three common award configurations: (i)~a single lump-sum award covering all applicants jointly; (ii)~differentiated per-applicant awards reflecting different roles in the underlying facts (for example, direct victim versus relative, detained person versus family member); and (iii)~mixed configurations in which some applicants receive an award and others receive no award for reasons such as \texttt{no\_claim}, \texttt{rule\_60\_non\_compliance}, or \texttt{finding\_sufficient}. 

\paragraph{Multi-violation cases.} A single judgment may also contain multiple violation findings. NPD award configurations take two forms: (i)~a single global NPD covering all violations and limbs, which is the most common configuration where violations are factually intertwined (for example, Article~2 substantive and procedural limbs arising from the same death); and (ii)~separated awards per violation, which is less common and typically used where violations are factually distinct.

\paragraph{Joined applications.}
The Court may process multiple applications together under \textit{Rule~42} of the Rules of Court~\citeyearpar{echr2025rules} when they raise related issues. Such applications may be decided in a single judgment and may share a common facts section or parallel factual sections. We do not divide a joined judgment into application-level examples, since doing so could place material from the same judicial document in different partitions.

\paragraph{Repetitive cases.}
Repetitive cases are distinct from joined applications. They concern separate applications arising from the same or a closely related legal problem, and they remain separate observations in the benchmark. \textit{Rule~61} permits the Court to initiate a pilot-judgment procedure where an application reveals a structural or systemic problem that has given rise, or may give rise, to similar applications~\citep{echr2025rules,echr2023pilotjudgment}. 

\paragraph{The Challenging diagnostic view.}
\label{app:challenging_set_define}
The Challenging view is a predefined, overlapping diagnostic subset based on criteria identified by a legal expert. It includes Grand Chamber judgments. From a legal-institutional perspective, these judgments often address legal questions of particular importance or complexity, warranting consideration by the Grand Chamber. It also includes cases that combine multiple applicants with multiple concurrent Convention violations. Targets for these cases are often holistic case-level NPD awards rather than a transparent sum of applicant-level awards or violation-level harms. Prediction for these cases requires one amount to be inferred from several potentially intertwined sources of harm, without assuming that applicant count or violation count maps linearly onto the award. We use this view to
test whether these expert-identified characteristics are associated with
greater prediction error; it does not attribute any observed difference to
a single cause.

\subsection{Equitable Assessment and Implicit Award Patterns}
\label{app:equity}

Article~41 allows the Court to award just satisfaction ``if necessary'', and the Practice Direction~\citeyearpar{echr2022justsatisfaction} confirms that non-pecuniary damage is assessed on an equitable basis rather than by precise calculation. There is no statutory formula for calculating the NPD, no published schedule of standard amounts, and no requirement that the Court give detailed reasons for the specific figure chosen in a given case. The benchmark therefore treats NPD as a continuous monetary target whose amount is estimated from permitted case information. Point-error metrics measure fit to the Court’s realised practice, not normative legal correctness.

\section{ECtHR-NPD Construction and Diagnostics}
\label{app:dataset-details}

\begin{figure*}[!htbp]
    \centering
    \includegraphics[width=\textwidth]{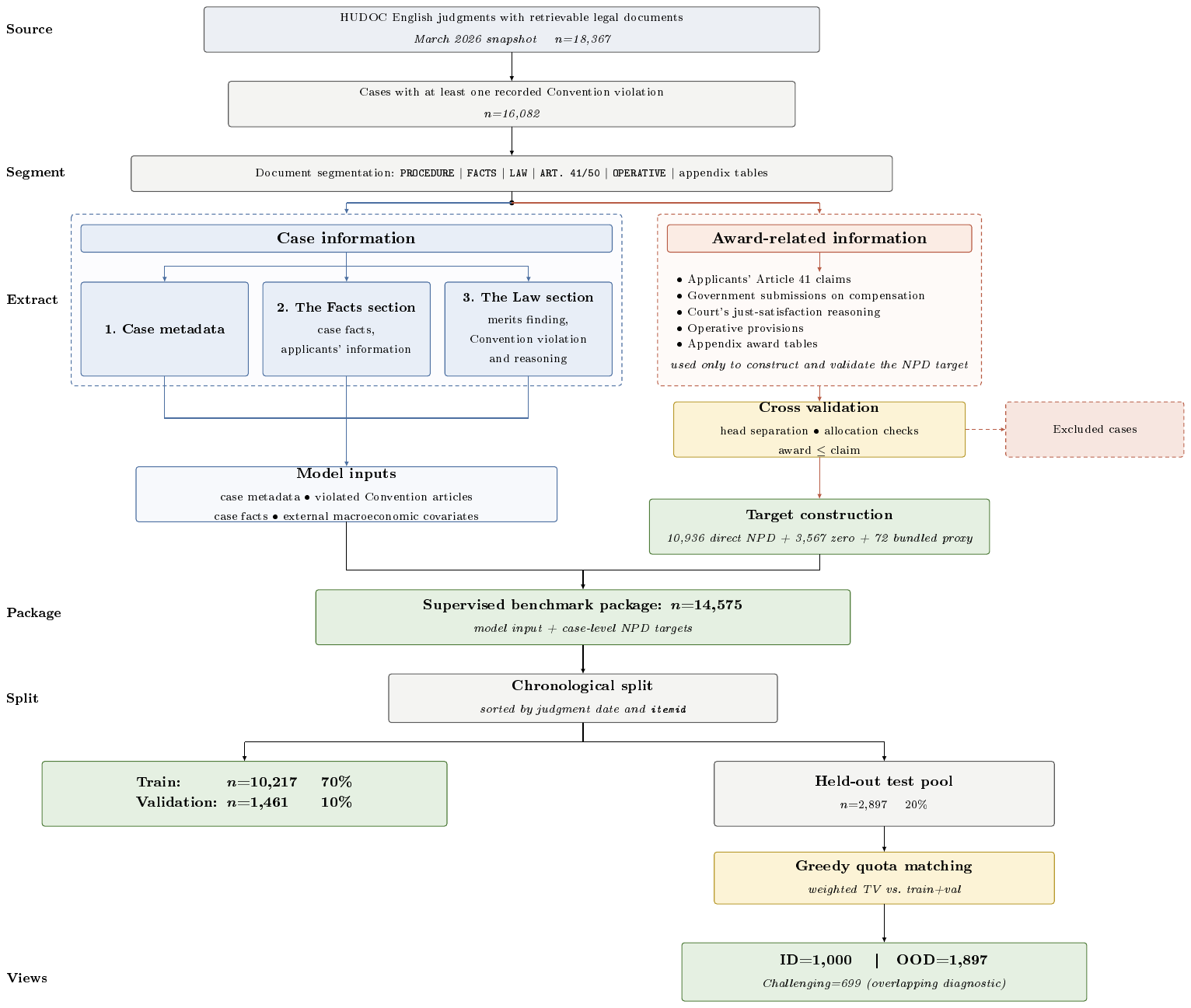}
    \caption{ECtHR-NPD construction pipeline. Source HUDOC judgments are segmented into permitted case information and award-related target-construction material. Model inputs exclude Article~41, operative awards, claim amounts, and target-related fields. Targets must pass checks for consistency. The dataset is partitioned into train/validation/test sets. Cases in the ID view are selected by a greedy quota matching algorithm; the OOD view contains the remaining test cases, and the Challenging view is an overlapping diagnostic subset.}
    \label{fig:ecthr-npd-pipeline}
\end{figure*}

This appendix reports the dataset construction, target validation, and diagnostic view splits for ECtHR-NPD. Figure~\ref{fig:ecthr-npd-pipeline} provides an overview of the dataset construction.

\subsection{Source Corpus and Exclusions}
\label{app:source-corpus}

We begin with 18{,}367 English HUDOC judgments with retrievable text (March~2026 snapshot). We apply the validation checks described in \S\ref{app:target-construction} and obtain a supervised pool of 14{,}575 cases. Table~\ref{tab:source-corpus-filtering-cascade} reports reasons for case exclusions and a cascade from the initial HUDOC retrieval to the final case pool. 

\begin{table}[!t]
\centering
\small
\setlength{\tabcolsep}{4pt}
\renewcommand{\arraystretch}{1.12}
\begin{tabular}{@{}>{\raggedright\arraybackslash}b{\dimexpr\linewidth-8.5em\relax}@{\hspace{0.75em}}r@{}}
\toprule
\textbf{Step} & \textbf{Cases retained} \\
\midrule
HUDOC English judgments with retrievable text
    & 18{,}367 \\
\hangindent=1.5em\hangafter=1\noindent
\quad $-$ no recorded Convention violation
    & 16{,}082 \\
\hangindent=1.5em\hangafter=1\noindent
\quad $-$ NPD amount not recoverable or not separable from bundled/multi-head award
    & 15{,}052 \\
\hangindent=1.5em\hangafter=1\noindent\quad $-$ no accepted zero/no-award rationale or sparse target evidence
    & 14{,}738 \\
\hangindent=1.5em\hangafter=1\noindent\quad $-$ historical Article~50 or pre-euro currency target unresolved
    & 14{,}575 \\
\midrule
\textbf{Supervised benchmark}
    & \textbf{14{,}575} \\
\bottomrule
\end{tabular}
\caption{Cascade of case exclusions from initial retrieved judgments from HUDOC to the final supervised pool.}
\label{tab:source-corpus-filtering-cascade}
\end{table}

\subsection{Data Sources and Extraction}
\label{app:field-routing-leakage-control}

The released data contain case metadata, structured case information, award-related material, merits findings, and external macroeconomic covariates. We separate these groups by whether they can be given to a model, are used to build or check the NPD target, or are kept for data checks and evaluation.

Some fields can be extracted directly without interpreting the judgment text. We obtain them from HUDOC metadata, judgment conclusion strings, accompanying release files, or external macroeconomic data matched by respondent state and judgment year. Other fields require interpretation of the judgment text. We extract these fields with prompting that returns structured fields and check them for consistency and leakage. Information used to build or check the target also undergoes the checks described in Section~\ref{app:target-construction}. We report where each group comes from and how it is used, rather than only counting fields obtained with fixed rules and fields obtained with a language model, because the release contains model inputs, case and applicant files, target records, and audit records with different purposes.

Table~\ref{tab:field-routing-leakage-control} shows these roles for each information group. ``Target Construction/Audit'' includes fields used to construct or validate the target, or to check extraction quality. These fields are not given to models for the target case.

\begin{table*}[!t]
\centering
\scriptsize
\setlength{\tabcolsep}{3pt}
\renewcommand{\arraystretch}{1.18}
\begin{tabularx}{\textwidth}{@{}
  >{\raggedright\arraybackslash}p{0.15\textwidth}
  >{\raggedright\arraybackslash}p{0.35\textwidth}
  >{\raggedright\arraybackslash}p{0.10\textwidth}
  >{\raggedright\arraybackslash}p{0.10\textwidth}
  >{\raggedright\arraybackslash}X@{}}
\toprule
\textbf{Segment / field group}
& \textbf{Information recorded}
& \textbf{Model Input}
& \textbf{Target Construction/Audit}
& \textbf{How the information is used} \\
\midrule

A Case metadata
& Public HUDOC identifiers, application numbers, judgment date, respondent state, court formation, case importance, and conclusion-derived violated articles.
& Selected fields
& Yes
& Case metadata and merits findings are given to models. Identifiers are retained for joining, splitting, and evaluation, but removed before model fitting. \\

B Applicant information and case facts
& Applicant counts, joined-application structure, representation status, non-identifying applicant aggregates, and facts-side case descriptors.
& Yes
& No
& Released in case and applicant files after identifying text is removed. These fields describe the case outside the Court's just-satisfaction reasoning.  \\

C Award-related material
& Applicants' Article~41 claims, Government submissions on compensation, NPD awards, per-applicant or beneficiary allocations, bundled-award flags, zero-award rationales, the Court's just-satisfaction reasoning, operative payment provisions, and appendix award tables.
& No
& Yes
& Used to build and check the NPD target. The ReAct result that adds applicants' claims and Article~41 reasoning is also reported separately from the main comparison.  \\

D Merits reasoning
& Violated Convention articles, number and type of violations, merits-side duration or severity factors, and other legal findings.
& Selected fields
& No
& Selected fields may be given to models after information about claims and awards is removed. The task conditions on the Court's merits outcome, but not its just-satisfaction reasoning. \\

E Other case factors and macroeconomic covariates
& Separate-opinion indicators, compact reasoning factors, and respondent-state macro covariates such as GDP and GDP per capita by judgment year.
&  Selected fields
& No
& Selected fields may be given to models after leakage checks. External covariates are matched by respondent state and judgment year. \\

\bottomrule
\end{tabularx}
\caption{Information used as model input and information used to build or check the NPD target. Model inputs exclude claims, awards, Article~41 text, operative provisions, appendix award tables, fields created while building the target, and evaluation records.}
\label{tab:field-routing-leakage-control}
\end{table*}

\subsection{Target Construction and Validation}
\label{app:target-construction}

The released target field \texttt{y\_amount\_eur} records the case-level NPD amount. We construct and validate it from three award-related sources: (1) a prompted LM extractor over the Article~41 section, (2) a structured cross-validator over the operative provisions, and (3) the appendix award tables. These sources are used only for target construction and validation. We apply eight validation checks: head separation, applicant-beneficiary consistency, per-applicant sum consistency, no-claim/positive-award incompatibility, claim-award consistency, currency normalisation, operative-provision recoverability, and a manually audited bundled-award exception. Retained targets are the nominal euro amounts. Historical Article~50 material is processed under the same logic only where it yields a recoverable NPD award; cases with unresolved Article~50 or pre-euro evidence are excluded. Cases that fail in any of these checks are manually reviewed or excluded.

\paragraph{Head separation.} The award must be non-pecuniary only. Bundled totals (where the operative provision awards a single sum spanning non-pecuniary damage, pecuniary damage, and costs) are excluded unless the bundled-proxy exception applies.

\paragraph{Bundled-proxy exception.} 230 bundled-award candidates were manually audited. 72 cases where the only valid claim head was non-pecuniary (no pecuniary or costs evidence anywhere in the judgment) enter the pool. The remaining 158 candidates carry other-head evidence and are excluded.

\paragraph{Zero-award cases.} Appendix~\ref{app:taxonomy} defines the legal taxonomy. Table~\ref{tab:zero-award-operational-counts} reports the distribution of the zero cases.

\begin{table}[h]
\centering
\small
\setlength{\tabcolsep}{5pt}
\renewcommand{\arraystretch}{1.15}
\begin{tabularx}{\linewidth}{@{}>{\raggedright\arraybackslash}X >{\raggedleft\arraybackslash}p{0.10\linewidth} >{\raggedleft\arraybackslash}p{0.20\linewidth}@{}}
\toprule
\textbf{Rationale} & \textbf{$n$} & \textbf{\% of zeros} \\
\midrule
\texttt{finding\_sufficient} & 2{,}075 & 58.2 \\
\texttt{no\_claim} & 1{,}287 & 36.1 \\
\texttt{unsubstantiated} & 124 & 3.5 \\
\texttt{rule\_60\_non\_compliance} & 60 & 1.7 \\
\texttt{domestic\_award\_covers} & 18 & 0.5 \\
\texttt{applicant\_deceased\_no\_heir} & 3 & 0.1 \\
\bottomrule
\end{tabularx}
\caption{Distribution of zero-award rationales.}
\label{tab:zero-award-operational-counts}
\end{table}

\subsection{Dataset Splits and Diagnostic Views}
\label{app:splits-and-diag-views}

\paragraph{Chronological split.} Cases are sorted by judgment date, with HUDOC item identifier as the tiebreaker, and partitioned 70/10/20 into train, validation, and test. This chronological partition prevents temporal leakage by keeping later judgments out of the data used to fit and select models, and reserves them for temporal generalisation. Resulting date ranges and target distributions are reported in Table~\ref{tab:target-distribution-by-split}.

\begin{table*}[t]
\centering
\scriptsize
\setlength{\tabcolsep}{3pt}
\renewcommand{\arraystretch}{1.15}
\begin{tabularx}{\textwidth}{@{}>{\raggedright\arraybackslash}p{0.12\textwidth} >{\raggedleft\arraybackslash}p{0.08\textwidth} >{\centering\arraybackslash}p{0.17\textwidth} >{\raggedleft\arraybackslash}p{0.08\textwidth} >{\raggedleft\arraybackslash}p{0.09\textwidth} >{\raggedleft\arraybackslash}p{0.11\textwidth} >{\raggedleft\arraybackslash}p{0.10\textwidth} >{\raggedleft\arraybackslash}p{0.11\textwidth}@{}}
\toprule
\textbf{Split} & \textbf{$n$} & \textbf{Date range} & \textbf{Zero $n$} & \textbf{Zero \%} & \textbf{Pos. median} & \textbf{Pos. p95} & \textbf{Max} \\
\midrule
Train & 10{,}217 & 1968-06-27 / 2019-07-18 & 2{,}248 & 22.0 & 5{,}000 & 46{,}480 & 10{,}697{,}900 \\
Validation & 1{,}461 & 2019-07-23 / 2021-12-16 & 366 & 25.1 & 7{,}500 & 89{,}900 & 1{,}887{,}300 \\
Test & 2{,}897 & 2021-12-16 / 2026-03-26 & 953 & 32.9 & 6{,}500 & 60{,}940 & 1{,}623{,}500 \\
\bottomrule
\end{tabularx}
\caption{Target distribution by chronological split. The validation and test ranges share the boundary date 16 December 2021; cases on that date are assigned deterministically by HUDOC item identifier. Amounts are in nominal euros.}
\label{tab:target-distribution-by-split}
\end{table*}

\paragraph{Diagnostic views.} The test pool is partitioned by greedy quota matching against the train plus validation reference distribution (Algorithm~\ref{alg:greedy-matching}) into ID (n=1{,}000) and OOD (n=1{,}897). The Challenging view (n=699) is an overlapping diagnostic view containing all Grand Chamber cases in the test period, together with cases that combine multiple applicants with multiple violations. 

For each matching dimension $d$, let
$p_{d,c}^{\mathrm{ref}}$ and $p_{d,c}^{v}$ denote the proportions of
category $c$ in the train-plus-validation reference distribution and
view $v$, respectively. We compute
\[
\operatorname{TV}_d(v)
=
\frac{1}{2}
\sum_{c \in \mathcal{C}_d}
\left|
p_{d,c}^{\mathrm{ref}} - p_{d,c}^{v}
\right|.
\]
The Mean TV reported in Tables~\ref{tab:splits-views}
and~\ref{tab:matching-diagnostics} is the unweighted mean of
$\operatorname{TV}_d(v)$ across the eight matching dimensions.
Court formation is reported as a diagnostic only and is excluded from
Mean TV. Tables~\ref{tab:diagnostic-views} and~\ref{tab:matching-diagnostics} report the diagnostic views, per-dimension TV, selection weights, and Mean TV.

\begin{algorithm}[t]
\small
\begin{algorithmic}[1]
\Require Reference set $R$ (train+val), test pool $T$, target size $k=1000$, dimensions $D$ with weights $w_d$
\Ensure ID partition $S\subset T$ and OOD partition $T\setminus S$

\State $q_{d,c} \gets |\{r\in R : r_d=c\}| \cdot k/|R|$ for all $d\in D$ and categories $c$
\State $S \gets \emptyset$; $n_{d,c} \gets 0$ for all $d,c$

\For{$i = 1,\ldots,k$}
    \ForAll{$t \in T\setminus S$}
        \State $\mathrm{score}(t) \gets
        \sum_{d\in D} w_d \cdot \max(0, q_{d,t_d}-n_{d,t_d})$
    \EndFor
    \State $t^* \gets \arg\max_{t\in T\setminus S}\mathrm{score}(t)$
    \State $S \gets S \cup \{t^*\}$
    \ForAll{$d\in D$}
        \State $n_{d,t^*_d} \gets n_{d,t^*_d}+1$
    \EndFor
\EndFor

\State \Return $S,\ T\setminus S$
\end{algorithmic}
\caption{Greedy quota matching for ID/OOD split}
\label{alg:greedy-matching}
\end{algorithm}

\begin{table*}[!t]
\centering
\scriptsize
\setlength{\tabcolsep}{4pt}
\renewcommand{\arraystretch}{1.15}
\begin{tabularx}{\textwidth}{@{}>{\raggedright\arraybackslash}p{0.12\textwidth} >{\raggedleft\arraybackslash}p{0.07\textwidth} >{\raggedright\arraybackslash}X >{\raggedleft\arraybackslash}p{0.10\textwidth} >{\raggedleft\arraybackslash}p{0.12\textwidth} >{\raggedleft\arraybackslash}p{0.10\textwidth} >{\raggedleft\arraybackslash}p{0.10\textwidth}@{}}
\toprule
\textbf{View} & \textbf{$n$} & \textbf{Definition} & \textbf{Zero \%} & \textbf{Pos. median} & \textbf{Pos. p95} & \textbf{Max} \\
\midrule
ID & 1{,}000 & Quota-matched subset of test pool against train+val reference. & 34.2 & 6{,}000 & 40{,}105 & 874{,}000 \\
OOD & 1{,}897 & Residual test cases not selected into ID. & 32.2 & 7{,}500 & 67{,}925 & 1{,}623{,}500 \\
Challenging & 699 & Grand Chamber judgments + cases with multi-applicant and multi-violation. & 46.1 & 15{,}000 & 144{,}478 & 1{,}623{,}500 \\
\bottomrule
\end{tabularx}
\caption{Diagnostic test views and their award distributions. ID and OOD partition the test pool; Challenging is an overlapping subset.}
\label{tab:diagnostic-views}
\end{table*}

\begin{table}[!t]
\centering
\small
\setlength{\tabcolsep}{5pt}
\renewcommand{\arraystretch}{1.15}
\begin{tabularx}{\linewidth}{@{}>{\raggedright\arraybackslash}X >{\raggedleft\arraybackslash}p{0.07\linewidth} >{\raggedleft\arraybackslash}p{0.09\linewidth} >{\raggedleft\arraybackslash}p{0.09\linewidth} >{\raggedleft\arraybackslash}p{0.13\linewidth}@{}}
\toprule
\textbf{Dimension} & \textbf{$w_d$} & \textbf{ID} & \textbf{OOD} & \textbf{Chal.} \\
\midrule
Respondent country & 6 & 0.065 & 0.305 & 0.484 \\
Case importance & 5 & 0.048 & 0.120 & 0.198 \\
Violation type & 3 & 0.009 & 0.036 & 0.218 \\
Applicant-count bucket & 3 & 0.007 & 0.389 & 0.854 \\
Violation-count bucket & 3 & 0.006 & 0.210 & 0.667 \\
Representation status & 2 & 0.007 & 0.099 & 0.019 \\
Separate-opinion flag & 2 & 0.008 & 0.093 & 0.090 \\
Top-20 violated articles & 2 & 0.051 & 0.433 & 0.949 \\
\midrule
Mean TV & & 0.025 & 0.211 & 0.435 \\
Court formation (diagnostic) & & 0.440 & 0.531 & 0.641 \\
\bottomrule
\end{tabularx}
\caption{Total variation (TV) of each matching dimension between the train plus validation reference and the diagnostic views. Lower is closer to the reference. The column $w_d$ gives the dimension weight used in the greedy selection algorithm. Court formation is reported as diagnostic only.}
\label{tab:matching-diagnostics}
\end{table}

\subsection{Structural Composition by Split}
\label{app:structural-composition}

Table~\ref{tab:structural-composition} reports the per-split distribution across the structural dimensions most relevant to the analyses in \S6: court formation, case importance, applicant-count bucket, violation-count bucket, and award bin. Per-split breakdowns of violated articles and respondent states are released with the dataset.

\begin{table*}[!t]
\centering
\small
\setlength{\tabcolsep}{5pt}
\renewcommand{\arraystretch}{1.12}
\begin{tabularx}{\textwidth}{@{}>{\raggedright\arraybackslash}p{0.24\textwidth} >{\raggedright\arraybackslash}X >{\raggedleft\arraybackslash}p{0.17\textwidth} >{\raggedleft\arraybackslash}p{0.17\textwidth} >{\raggedleft\arraybackslash}p{0.17\textwidth}@{}}
\toprule
\textbf{Dimension} & \textbf{Category} & \textbf{Train} & \textbf{Val} & \textbf{Test} \\
\midrule
Court formation & Chamber & 8{,}008 (78.4) & 493 (33.7) & 706 (24.4) \\
& Committee & 1{,}986 (19.4) & 953 (65.2) & 2{,}176 (75.1) \\
& Grand Chamber & 223 (2.2) & 15 (1.0) & 15 (0.5) \\
\addlinespace
Case importance & 1 & 572 (5.6) & 52 (3.6) & 45 (1.6) \\
& 2 & 445 (4.4) & 10 (0.7) & 0 (0.0) \\
& 3 & 2{,}104 (20.6) & 261 (17.9) & 544 (18.8) \\
& 4 & 7{,}096 (69.5) & 1{,}138 (77.9) & 2{,}308 (79.7) \\
\addlinespace
Applicant-count bucket & 1 & 7{,}755 (75.9) & 933 (63.9) & 1{,}569 (54.2) \\
& 2--5 & 1{,}852 (18.1) & 369 (25.3) & 768 (26.5) \\
& 6--20 & 535 (5.2) & 132 (9.0) & 412 (14.2) \\
& 21--100 & 68 (0.7) & 24 (1.6) & 139 (4.8) \\
& $>$100 & 7 (0.1) & 3 (0.2) & 9 (0.3) \\
\addlinespace
Violation-count bucket & 1 & 6{,}748 (66.0) & 931 (63.7) & 1{,}753 (60.5) \\
& 2 & 2{,}258 (22.1) & 253 (17.3) & 371 (12.8) \\
& 3--5 & 1{,}112 (10.9) & 260 (17.8) & 680 (23.5) \\
& $>$5 & 99 (1.0) & 17 (1.2) & 93 (3.2) \\
\addlinespace
Award bin & 0 & 2{,}248 (22.0) & 366 (25.1) & 953 (32.9) \\
& $>$0--10k & 6{,}025 (59.0) & 729 (49.9) & 1{,}333 (46.0) \\
& 10k--50k & 1{,}592 (15.6) & 274 (18.8) & 493 (17.0) \\
& $>$50k & 352 (3.4) & 92 (6.3) & 118 (4.1) \\
\bottomrule
\end{tabularx}
\caption{Structural composition diagnostics by split. Cell entries are $n\ (\%)$.}
\label{tab:structural-composition}
\end{table*}

\section{Experimental Design and Setup}
\label{app:experimental-design-setup}

This appendix records the operational settings for the experiments in Section~\ref{sec:experiments}. Every system predicts one case-level amount (\texttt{award\_eur}) in nominal euros for Article~41 non-pecuniary damage, with zero as a valid target. The common protocol makes the evaluation comparable across different methods. Input settings that differ from the benchmark model input are reported as diagnostic conditions rather than folded into the baseline comparison.

\subsection{Common Protocol}
\label{app:common-protocol}

\begin{table}[!t]
\centering
\small
\begin{tabularx}{\linewidth}{@{}p{0.32\linewidth}X@{}}
\toprule
\textbf{Field} & \textbf{Setting} \\
\midrule
Task & Case-level continuous monetary prediction \\
Target & \texttt{y\_amount\_eur} in nominal euros \\
Output & One non-negative number per case \\
Cases & 10{,}217 train, 1{,}461 validation, 2{,}897 test \\
Split policy & Chronological 70/10/20, judgment date then \texttt{itemid} \\
Test views & ID (1{,}000) and OOD (1{,}897) partition the test pool; Challenging (699) is an overlapping diagnostic \\
Primary metric & MAE in linear EUR \\
Diagnostic metrics & RMSE, $R^2$, MedAE, 95AE, Pearson~$r$, Spearman~$\rho$, Zero-F1, bucket MAE \\
Model selection & Validation split only \\
\bottomrule
\end{tabularx}
\caption{Common protocol shared by all systems.}
\label{tab:common-protocol}
\end{table}

\paragraph{Model input setting.}
Models do not receive Article~41 claims submitted by the target case’s applicants, Government submissions on compensation, the Court’s just-satisfaction reasoning, operative provisions, or appendix award tables. They also do not receive direct award snippets, claim-state or no-claim fields, zero-award-rationale labels, raw extractor outputs, or per-applicant award allocations. Few-shot prompting with retrieved examples (Section~\ref{app:fewshot}) may include awards and award rationales from training references. The Expanded ReAct diagnostic (Section~\ref{app:react-retrieval}) adds target case claim and Article 41 reasoning to the model input, but it continues to withhold the final award and structured target-derived fields.

\subsection{Constant Baselines}
\label{app:constants}

\begin{table}[!t]
\centering
\small
\begin{tabularx}{\linewidth}{@{}p{0.32\linewidth}X@{}}
\toprule
\textbf{Field} & \textbf{Setting} \\
\midrule
Predictors & Training-set median, training-set mean, constant zero (plus diagnostic positive-only median and mean) \\
Input & None \\
Estimation & Training-split labels only \\
Evaluation scale & Nominal EUR \\
\bottomrule
\end{tabularx}
\caption{Constant baselines (sanity floors).}
\label{tab:setup-constants}
\end{table}

Table~\ref{tab:constant-baseline-values} reports the constant predictors and their view-level MAE.

\begin{table*}[!t]
\centering
\small
\setlength{\tabcolsep}{5pt}
\begin{tabular}{lrrrrrr}
\toprule
\textbf{Constant} & \textbf{Value} & \textbf{Val MAE} & \textbf{Test MAE} & \textbf{ID MAE} & \textbf{OOD MAE} & \textbf{Chal. MAE} \\
\midrule
Constant zero               &      0 & 21{,}257 & 11{,}713 & 10{,}309 & 12{,}454 & 22{,}916 \\
Train median                & 3{,}200 & 20{,}145 & 11{,}006 &  9{,}752 & 11{,}668 & 22{,}812 \\
Train mean                  & 13{,}010 & 22{,}944 & 14{,}940 & 14{,}465 & 15{,}190 & 25{,}150 \\
Train positive median       & 5{,}000 & 20{,}087 & 11{,}173 & 10{,}062 & 11{,}760 & 22{,}948 \\
Train positive mean         & 16{,}680 & 25{,}125 & 17{,}448 & 17{,}214 & 17{,}572 & 26{,}846 \\
\bottomrule
\end{tabular}
\caption{Constant baselines and their MAEs in different views; amounts are in EUR.}
\label{tab:constant-baseline-values}
\end{table*}

\subsection{Gradient-boosted Trees}
\label{app:trees}

\begin{table}[!t]
\centering
\small
\begin{tabularx}{\linewidth}{@{}p{0.32\linewidth}X@{}}
\toprule
\textbf{Field} & \textbf{Setting} \\
\midrule
Models & CatBoost, XGBoost, LightGBM \\
Input & Leakage-controlled structured features: case metadata, violated-article indicators, applicant aggregates, merits reasoning aggregates, and respondent-state macro covariates \\
Target transform & $\log(1+y)$ during training \\
Prediction transform & $\exp(\hat{y}_{\log})-1$, clipped at zero \\
Tuning & Validation MAE; median AE as tie-breaker \\
Numeric handling & Median imputation from train \\
Categorical handling & One-hot with unknown-category handling (CatBoost uses native categorical pools) \\
\bottomrule
\end{tabularx}
\caption{Tree-regression settings.}
\label{tab:setup-trees}
\end{table}

All claim-related, award-related, and target-derived columns are dropped
before fitting. The case identifier (\texttt{itemid}) is dropped to
prevent memorisation and is retained only for alignment and output
joins.

\subsection{Retrieval Baselines}
\label{app:retrieval}

\begin{table}[!t]
\centering
\small
\begin{tabularx}{\linewidth}{@{}p{0.32\linewidth}X@{}}
\toprule
\textbf{Field} & \textbf{Setting} \\
\midrule
Retrievers & $k$-NN over structured features; BM25 (\texttt{bm25s}~0.3.8) over the \textsc{Facts} section; BGE-M3 dense and sparse over the \textsc{Facts} section \\
Top-$k$ & 20 (selected on validation) \\
Aggregation & Median award across retrieved training neighbours \\
Temporal filter & Reference judgment date strictly precedes target judgment date \\
Article filter & At least one shared violated article \\
Target exclusion & Reference \texttt{itemid} cannot equal target \texttt{itemid} \\
Fallback & Training-set median when no eligible neighbours exist \\
\bottomrule
\end{tabularx}
\caption{Retrieval baseline settings. Retrieval pool and label source are restricted to the training set.}
\label{tab:setup-retrieval}
\end{table}

\subsection{Encoder LMs}
\label{app:encoders}

\begin{table}[!t]
\centering
\small
\begin{tabularx}{\linewidth}{@{}p{0.32\linewidth}X@{}}
\toprule
\textbf{Field} & \textbf{Setting} \\
\midrule
Models & ModernBERT-base (8{,}192 tokens); Legal-Longformer (4{,}096 tokens, LegalBERT-initialised) \\
Input & the \textsc{Facts} section \\
Long-document strategy & ModernBERT: single 8{,}192-token forward pass. Legal-Longformer: two 4{,}096-token chunks with mean pooling over chunk-level \texttt{[CLS]} \\
Output head & Linear regression over \texttt{[CLS]} \\
Loss & SmoothL1 ($\beta=1.0$) on standardised $\log(1+y)$ \\
Prediction transform & Inverse-standardise, $\exp(\hat{z})-1$, clip at zero \\
Optimiser & AdamW, weight decay 0.01, linear warmup over 6\% of steps, linear decay \\
Precision and batch & bf16; effective batch size 16 (batch 1, grad-accum 16); gradient checkpointing \\
Late-fusion variant & Concatenate \texttt{[CLS]} with a structured MLP branch (hidden 256, output 768) before the regression head \\
\bottomrule
\end{tabularx}
\caption{Encoder and late-fusion settings.}
\label{tab:setup-encoders}
\end{table}

\subsection{Prompted LMs}
\label{app:llm}

\begin{table}[!t]
\centering
\small
\begin{tabularx}{\linewidth}{@{}p{0.32\linewidth}X@{}}
\toprule
\textbf{Field} & \textbf{Setting} \\
\midrule
Conditions & Vanilla zero-shot; static chain-of-thought; retrieved few-shot CoT \\
Models & Qwen3.5-9B, Qwen3.5-27B, Qwen3.5-Plus, GPT-OSS-20B, GPT-5.4 \\
Model records & Provider, model snapshot, access date, and decoding settings are recorded in the run manifest \\
Input & case metadata, the \textsc{Facts} section (serialised extracted features from this section for the structured input setting), violated Articles, and external macroeconomic covariates \\
Temperature & 0 where supported; provider low-temperature default otherwise \\
Seed & 42 where supported \\
Max output tokens & 4{,}096 \\
JSON handling & Provider JSON schema when stable; otherwise schema embedded in prompt with local validation \\
External tools & Disabled in all prompted conditions \\
\bottomrule
\end{tabularx}
\caption{Prompted-LM common settings.}
\label{tab:setup-llm}
\end{table}

\paragraph{Vanilla zero-shot.}
The prompt asks the model for one case-level award given the model input. No reasoning scaffold, examples, retrieval, or knowledge-base access is offered. The output is
a JSON object containing only \texttt{award\_eur}.

\paragraph{Static chain-of-thought.}
The static-CoT prompt guides the model through five steps: identifying case factors, assessing the possibility of a zero award, evaluating damages, considering applicant aggregation, and calibrating the award amount. It then asks for a brief rationale and a single case-level prediction. Zero is treated as a valid value of the continuous target, and the model is instructed not to sum awards across Articles or multiply them by the number of applicants. External retrieval and award cues from named cases, citations, application numbers, or dataset-level prevalence are prohibited.

\paragraph{Retrieved few-shot CoT.}
\label{app:fewshot}
The target case retains the common model input. The prompt is augmented with up to
five training references retrieved under the same temporal and article
filters as the retrieval baselines. Candidates are ranked by a
weighted combination of exact article match, article-overlap
Jaccard, respondent state, violation type, applicant-count band, court
formation, case importance, and recency. The selected set contains the highest-ranked positive-award cases and one zero-award case when available. The prompt includes each reference award and its Article~41 reasoning. Because this condition exposes labelled training examples that are unavailable to zero-shot and static-CoT prompting, we report it separately.

\paragraph{Serialised input variant.}
\label{app:serialization}
For vanilla and static-CoT prompting, we also replace the case facts with a text serialisation of the same structured features used by the tree models. This both reduces the amount of text the LM must process and allows comparison with tree models using the same input information.

\subsection{Knowledge-Augmented ReAct Agents}
\label{app:react}

The ReAct configuration is a knowledge-augmented diagnostic setting. The controller
mediates every model action through a whitelisted tool set. The model
never reads judgment files directly; the controller redacts each tool
observation before returning it and applies a leakage gate before the
final prediction is scored. In the base diagnostic, the target case follows the common model input setting. 

\begin{table}[!t]
\centering
\small
\begin{tabularx}{\linewidth}{@{}p{0.32\linewidth}X@{}}
\toprule
\textbf{Field} & \textbf{Setting} \\
\midrule
Models & Qwen3.5-Plus (DashScope), MiniMax-M2.7 (MiniMax API) \\
Execution & Bounded ReAct, JSON action objects \\
Search budget & Up to 12 steps, top-5 retrieved references \\
Temperature & Provider deterministic or near-deterministic setting recorded in run manifest \\
Trace & Per-case log of actions, observations, retrievals, priors, leakage checks, final prediction \\
Final output & \texttt{final\_predict} returning \texttt{award\_eur} plus diagnostic fields (rationale summary, zero/positive decision, aggregation-scale decision, uncertainty) \\
\bottomrule
\end{tabularx}
\caption{ReAct controller settings.}
\label{tab:react-settings}
\end{table}

\paragraph{Controller actions.}

\begin{table}[!htbp]
\centering
\small
\setlength{\tabcolsep}{4pt}
\renewcommand{\arraystretch}{1.08}
\begin{tabularx}{\linewidth}{@{}
  >{\raggedright\arraybackslash}p{0.38\linewidth}
  >{\raggedright\arraybackslash}X@{}}
\toprule
\textbf{Action} & \textbf{Function} \\
\midrule
\texttt{inspect\_case}
& Read the redacted target-case overview. \\

\shortstack[l]{\texttt{query\_target}\\\texttt{information}}
& Query allowed structured target features. \\

\shortstack[l]{\texttt{search\_modules},\\\texttt{load\_module}}
& Find and load knowledge-base modules. \\

\shortstack[l]{\texttt{resolve\_empirical}\\\texttt{priors}}
& Retrieve train-only article, country, and article--country priors. \\

\shortstack[l]{\texttt{retrieve\_train}\\\texttt{references}}
& Retrieve temporally prior training cases. \\

\shortstack[l]{\texttt{query\_reference}\\\texttt{features}}
& Inspect retrieved-reference features; award-derived fields are excluded. \\

\shortstack[l]{\texttt{assess\_aggregation}\\\texttt{pattern}}
& Summarise multi-applicant or multi-violation aggregation before prediction. \\

\texttt{leakage\_check}
& Run a pre-prediction redaction audit. \\

\texttt{final\_predict}
& Emit the final JSON prediction. \\
\bottomrule
\end{tabularx}
\caption{Whitelisted ReAct controller actions.}
\label{tab:react-actions}
\end{table}

\paragraph{Reference retrieval.}
\label{app:react-retrieval}
Retrieval is restricted to the training split and applies the
same temporal and article filters as the retrieval baselines.
Candidates are ranked by exact article-set match, article-overlap
Jaccard, respondent state, violation type, applicant-count band, court
formation, case importance, and recency. The controller may
balance the training-reference pool across positive and zero observed training targets before selecting similarity-ranked reference cases. Empirical priors
follow the fallback order article-by-country, article, country, then
a global article-weighted prior.

\paragraph{Knowledge base.}
The knowledge base is organised into four module families:
\textit{normative} modules covering legal criteria, severity drivers,
and aggregation rules for each Convention article in scope, plus
cross-article synthesis, claim rules, finding-sufficient guidance, and
zero-award classification; \textit{empirical} modules exposing the
train-only priors above with worked examples; \textit{routing} modules
including the article limb router; and \textit{policy} modules containing the system prompt, the
ReAct action protocol, the zero/non-zero gate, the output schema, and
the redaction contract. The controller enforces this
protocol at each step. Modules are loaded on
demand based on the target case's violated-article set.

\paragraph{Trace auditing.}
Each run writes per-case traces, predictions, action logs, tool observations, leakage-check outcomes, and final JSON output. These records support reproducibility and error analysis. They do not alter predictions or provide evidence that a generated rationale faithfully represents the model’s decision process.

\paragraph{Expanded ReAct diagnostic.}
\label{app:expanded-react}
The expanded ReAct diagnostic receives the target's NPD claim and Article~41 reasoning in addition to the information available to the base agent. The target's final numeric award, operative payment clause, award tables, and target-derived fields remain withheld. This is nevertheless not a benchmark-input condition: Article~41 reasoning may explain why the Court made no monetary award or treated the finding of a violation as sufficient just satisfaction. We report this condition separately to assess how much these award-related materials improve prediction, rather than as part of the main model comparison (Table~\ref{tab:agent-input-diagnostics}).

\section{Experiment Results and Diagnostic Views}
\label{app:experiment-results-diagnostic-views}

This appendix reports the additional analyses used to interpret the main results.
It compares raw and structured inputs, reports MAE on the ID, OOD, and
Challenging test views, breaks down errors by respondent state and violated
Article, and summarises prediction errors by award range. We also report
serialised input results for prompted LMs separately. Expanded ReAct results are shown only as
diagnostic analyses because these runs expose claim information and Article~41
reasoning that are excluded from the benchmark input.

\subsection{Paired Bootstrap Significance Tests}
\label{app:bootstrap-significance}
Table~\ref{tab:bootstrap-significance} reports paired bootstrap comparisons for selected system pairs to assess whether the main MAE differences are statistically reliable.

\begin{table*}[!t]
\centering
\small
\setlength{\tabcolsep}{4pt}
\renewcommand{\arraystretch}{1.05}
\resizebox{\textwidth}{!}{%
\begin{tabular}{llrrrrl}
\toprule
System $A$ & Comparison $B$ & MAE($A$) & MAE($B$) & $\Delta$ & $\Pr(A<B)$ & Result \\
\midrule
\multicolumn{7}{l}{\textit{Strict non-decoder and baseline comparisons}} \\
CatBoost & BGE-M3 dense & 9{,}881 & 10{,}177 & -296 & 0.935 & n.s. \\
CatBoost & BGE-M3 sparse & 9{,}881 & 10{,}226 & -345 & 0.996 & A better \\
CatBoost & ModernBERT late fusion & 9{,}881 & 10{,}074 & -193 & 0.807 & n.s. \\
CatBoost & Qwen3.5-9B static CoT & 9{,}881 & 10{,}782 & -901 & 1.000 & A better \\
CatBoost & Train median & 9{,}881 & 11{,}006 & -1{,}125 & 1.000 & A better \\
Qwen3.5-9B static CoT & Train median & 10{,}782 & 11{,}006 & -224 & 0.820 & n.s. \\
\midrule
\multicolumn{7}{l}{\textit{Decoder-LM prompting and clean agent comparisons}} \\
Static CoT (Qwen3.5-9B) & Vanilla zero-shot (GPT-OSS-20B) & 10{,}782 & 15{,}678 & -4{,}896 & 1.000 & A better \\
Few-shot CoT (GPT-OSS-20B)$^\dagger$ & Vanilla zero-shot (GPT-OSS-20B) & 11{,}674 & 15{,}678 & -4{,}004 & 1.000 & A better \\
Static CoT (Qwen3.5-9B) & Few-shot CoT (GPT-OSS-20B)$^\dagger$ & 10{,}782 & 11{,}674 & -892 & 0.906 & n.s. \\
Static CoT (Qwen3.5-9B) & base ReAct (MiniMax-M2.7)$^\dagger$ & 10{,}782 & 16{,}571 & -5{,}789 & 1.000 & A better \\
Few-shot CoT (GPT-OSS-20B)$^\dagger$ & base ReAct (MiniMax-M2.7)$^\dagger$ & 11{,}674 & 16{,}571 & -4{,}897 & 1.000 & A better \\
Vanilla zero-shot (GPT-OSS-20B) & base ReAct (MiniMax-M2.7)$^\dagger$ & 15{,}678 & 16{,}571 & -893 & 0.792 & n.s. \\
Few-shot CoT (GPT-OSS-20B)$^\dagger$ & Train median & 11{,}674 & 11{,}006 & 668 & 0.002 & B better \\
Vanilla zero-shot (GPT-OSS-20B) & Train median & 15{,}678 & 11{,}006 & 4{,}672 & 0.000 & B better \\
base ReAct (MiniMax-M2.7)$^\dagger$ & Train median & 16{,}571 & 11{,}006 & 5{,}565 & 0.000 & B better \\
\bottomrule
\end{tabular}
}
\caption{Paired bootstrap comparisons on test-pool MAE. For each system pair, MAE values and $\Delta=\mathrm{MAE}(A)-\mathrm{MAE}(B)$ are computed on the cases with scorable predictions from both systems. Each comparison uses 1{,}000 full-size paired bootstrap resamples of this shared set. $\Pr(\Delta^*<0)$ is the fraction of resamples in which system $A$ has lower MAE than system $B$; negative values of $\Delta$ therefore favour system $A$. Daggered rows are diagnostic settings.}
\label{tab:bootstrap-significance}
\end{table*}

\subsection{Ablation Results}
\label{app:ablation}

\begin{table}[H]
\centering
\small
\setlength{\tabcolsep}{5pt}
\renewcommand{\arraystretch}{1.05}
\begin{tabular}{lrrrrr}
\toprule
Input & Mean & Median & P95 & Max \\
\midrule
Raw text & 1{,}945 & 1{,}454 & 5{,}337 & 43{,}395 \\
Serialised features & 102 & 67 & 237 & 8{,}640 \\
\bottomrule
\end{tabular}
\caption{Input length comparison for raw-text and
serialised structured features on the test pool. Word counts are computed
on the model input, including the provided violated-article header.}
\label{tab:input-length-comparison}
\end{table}

This appendix reports the input-representation ablations referenced in
Section~\ref{sec:input-calibration-ablations}. Table~\ref{tab:tree-feature-ablation}
shows structured-feature tree ablations, and
Table~\ref{tab:encoder-fusion-ablation} shows text-only versus late-fusion
encoder results.

\begin{table*}[!t]
\centering
\small
\setlength{\tabcolsep}{4pt}
\renewcommand{\arraystretch}{1.05}
\begin{tabular}{lrrrrrr}
\toprule
Feature set
& \multicolumn{2}{c}{CatBoost}
& \multicolumn{2}{c}{XGBoost}
& \multicolumn{2}{c}{LightGBM} \\
\cmidrule(lr){2-3}
\cmidrule(lr){4-5}
\cmidrule(lr){6-7}
& Val & Test & Val & Test & Val & Test \\
\midrule
X0 case metadata
& 16{,}275 & 10{,}424
& 16{,}487 & 10{,}922
& 16{,}991 & 10{,}346 \\
X1 applicant-enhanced
& 14{,}795 & 9{,}881
& 15{,}892 & 10{,}117
& 15{,}627 & 10{,}240 \\
X2 reasoning-enhanced
& 16{,}413 & 10{,}276
& 16{,}854 & 11{,}086
& 17{,}040 & 11{,}316 \\
X3 full feature set
& 14{,}980 & 9{,}974
& 15{,}874 & 10{,}226
& 15{,}797 & 9{,}863 \\
\bottomrule
\end{tabular}
\caption{Structured-feature tree ablations by feature set. Values are
MAE in EUR. The headline tree results in Table~\ref{tab:main-results} use X1, the applicant-enhanced feature set. We fix X1 for all three tree models for cross-model comparability. X0--X3 vary the input features only; the prediction target is unchanged. For each model and feature set, hyperparameters are selected
by validation MAE; the corresponding validation and test MAE are reported.}
\label{tab:tree-feature-ablation}
\end{table*}

\begin{table}[t]
\centering
\small
\setlength{\tabcolsep}{4pt}
\renewcommand{\arraystretch}{1.05}
\resizebox{\columnwidth}{!}{%
\begin{tabular}{llrrrrr}
\toprule
Model & Input & MAE & $\Delta$ & ID & OOD & Chal. \\
\midrule
ModernBERT & raw text
& 10{,}133 & -- & 8{,}438 & 11{,}026 & 21{,}544 \\
ModernBERT & late fusion
& 10{,}074 & -59 & 8{,}404 & 10{,}955 & 21{,}024 \\
Legal-Longformer & raw text
& 10{,}642 & -- & 9{,}200 & 11{,}402 & 22{,}116 \\
Legal-Longformer & late fusion
& 10{,}244 & -398 & 8{,}695 & 11{,}060 & 21{,}013 \\
\bottomrule
\end{tabular}
}
\caption{Encoder input ablations. Late fusion concatenates the encoder
representation with leakage-audited structured features before the
regression head. $\Delta$ is MAE change relative to the text-only
variant of the same encoder; negative values indicate improvement.}
\label{tab:encoder-fusion-ablation}
\end{table}

\subsection{Diagnostic Views}
\label{app:diagnostic-test-views}

The three views answer different descriptive questions about the same held-out period. ID tests temporal generalisation where the observed case structure is closer to the train-plus-validation reference; OOD is the remaining part of the test pool; and Challenging reports performance for a pre-specified, overlapping composition of cases. A lower ID MAE is therefore consistent with closer observed composition, but does not establish that any individual legal characteristic causes lower error.

Across the representative systems in Table~\ref{tab:diagnostic-view-degradation}, error rises in the OOD and Challenging views. The pattern is useful for locating where aggregate MAE hides instability, especially under higher zero prevalence, different respondent-state mix, and multi-applicant or multi-violation case structure. It should be read as a diagnostic description of this benchmark split rather than as a causal account of legal difficulty.

\begin{table*}[!t]
\centering
\small
\setlength{\tabcolsep}{4pt}
\renewcommand{\arraystretch}{1.05}
\begin{tabular}{llrrrr}
\toprule
Representative system & Family & Test MAE & ID MAE & OOD MAE & Challenging MAE \\
\midrule
CatBoost & tree
& 9{,}881 & 8{,}143 ($\downarrow$17.6\%) & 10{,}797 ($\uparrow$9.3\%) & 20{,}987 ($\uparrow$112.4\%) \\
BGE-M3 sparse & retrieval
& 10{,}226 & 8{,}796 ($\downarrow$14.0\%) & 10{,}980 ($\uparrow$7.4\%) & 20{,}378 ($\uparrow$99.3\%) \\
ModernBERT late fusion & encoder
& 10{,}074 & 8{,}404 ($\downarrow$16.6\%) & 10{,}955 ($\uparrow$8.7\%) & 21{,}024 ($\uparrow$108.7\%) \\
Qwen3.5-9B CoT & prompted LM
& 10{,}782 & 9{,}777 ($\downarrow$9.3\%) & 11{,}312 ($\uparrow$4.9\%) & 20{,}926 ($\uparrow$94.1\%) \\
Qwen3.5-Plus zero-shot & frontier LM
& 16{,}400 & 12{,}781 ($\downarrow$22.1\%) & 18{,}306 ($\uparrow$11.6\%) & 37{,}186 ($\uparrow$126.7\%) \\
Qwen3.5-Plus few-shot$^\dagger$ & frontier LM
& 24{,}141 & 13{,}518 ($\downarrow$44.0\%) & 29{,}741 ($\uparrow$23.2\%) & 60{,}814 ($\uparrow$151.9\%) \\
MiniMax-M2.7 base ReAct$^\dagger$ & ReAct diagnostic
& 16{,}571 & 10{,}911 ($\downarrow$34.2\%) & 19{,}555 ($\uparrow$18.0\%) & 40{,}758 ($\uparrow$146.0\%) \\
\bottomrule
\end{tabular}
\caption{Degradation across diagnostic test views. MAE values are in EUR. Percentages show change relative to the full test-pool MAE for the same
system.}
\label{tab:diagnostic-view-degradation}
\end{table*}

\subsection{Results by Respondent State and Violated Articles}
\label{app:state-article-side-views}

Tables~\ref{tab:state-docket-side-view} and \ref{tab:article-side-view} report results by respondent state and violated Article for representative systems. These analyses are descriptive rather than comparative rankings. Article rows overlap because a single judgment may involve multiple violated provisions.

\begin{table*}[!htbp]
\centering
\small
\setlength{\tabcolsep}{3pt}
\renewcommand{\arraystretch}{1.05}
\resizebox{\textwidth}{!}{%
\begin{tabular}{lrrrrrrrr}
\toprule
Respondent group & $n$ & Zero & \shortstack{Multi-\\applicant} & Pos. p95
& \shortstack{Cat\\Boost} & \shortstack{ModernBERT\\LF}
& \shortstack{Qwen3.5-9B\\CoT}
& \shortstack{MiniMax-M2.7\\base ReAct$^\dagger$} \\
\midrule
Russian Federation & 743 & 54.2\% & 72.5\% & 158{,}425 & 18{,}585 & 18{,}605 & 19{,}620 & 40{,}015 \\
Ukraine & 494 & 18.8\% & 46.2\% & 53{,}100 & 8{,}540 & 8{,}895 & 7{,}695 & 7{,}361 \\
Turkey & 153 & 33.3\% & 35.9\% & 25{,}395 & 4{,}227 & 4{,}434 & 5{,}399 & 15{,}612 \\
Azerbaijan & 133 & 7.5\% & 39.1\% & 31{,}800 & 4{,}987 & 5{,}775 & 5{,}121 & 8{,}256 \\
Hungary & 124 & 51.6\% & 56.5\% & 25{,}050 & 3{,}830 & 4{,}560 & 7{,}035 & 10{,}067 \\
Romania & 103 & 35.9\% & 50.5\% & 111{,}250 & 14{,}264 & 12{,}393 & 18{,}418 & 20{,}197 \\
Poland & 103 & 31.1\% & 21.4\% & 29{,}650 & 7{,}197 & 5{,}666 & 6{,}764 & 6{,}490 \\
Italy & 98 & 27.6\% & 46.9\% & 41{,}600 & 8{,}855 & 9{,}296 & 10{,}267 & 11{,}655 \\
Moldova & 92 & 13.0\% & 19.6\% & 35{,}750 & 4{,}651 & 5{,}733 & 7{,}199 & 5{,}563 \\
Armenia & 89 & 9.0\% & 27.0\% & 39{,}000 & 6{,}446 & 7{,}519 & 5{,}389 & 5{,}564 \\
Croatia & 83 & 21.7\% & 12.0\% & 16{,}040 & 4{,}564 & 4{,}719 & 3{,}814 & 3{,}361 \\
Bulgaria & 68 & 25.0\% & 38.2\% & 22{,}500 & 5{,}792 & 5{,}776 & 5{,}622 & 5{,}733 \\
Other & 614 & 29.5\% & 30.3\% & 32{,}000 & 6{,}449 & 6{,}845 & 8{,}204 & 7{,}065 \\
\bottomrule
\end{tabular}
}
\caption{Respondent-state side view for representative systems. Rows show
the twelve largest respondent groups in the test pool plus all other
cases. Multi-applicant is the share of cases with \texttt{num\_applicants}
$>1$. MAE and positive-award p95 are in EUR. The rows are descriptive
docket slices, not estimates of a respondent-state effect.}
\label{tab:state-docket-side-view}
\end{table*}

\begin{table*}[!t]
\centering
\small
\setlength{\tabcolsep}{3pt}
\renewcommand{\arraystretch}{1.05}
\begin{tabular}{lrrrrrrrrr}
\toprule
Provision & $n$ & Zero & Multi-viol. & Pos. p95
& \shortstack{Cat\\Boost} & \shortstack{BGE-M3\\dense} & \shortstack{ModernBERT\\LF} & \shortstack{Qwen3.5-9B\\CoT} & \shortstack{MiniMax-M2.7\\base ReAct$^\dagger$} \\
\midrule
Art.~6 & 1{,}152 & 37.8\% & 56.4\% & 71{,}240 & 8{,}941 & 10{,}258 & 9{,}267 & 10{,}001 & 14{,}524 \\
Art.~5 & 757 & 37.9\% & 68.4\% & 99{,}275 & 14{,}059 & 14{,}107 & 13{,}984 & 15{,}811 & 26{,}349 \\
Art.~3 & 692 & 31.2\% & 65.2\% & 94{,}500 & 16{,}873 & 16{,}987 & 17{,}227 & 19{,}759 & 33{,}438 \\
Art.~8 & 499 & 37.1\% & 44.5\% & 58{,}375 & 10{,}917 & 11{,}204 & 11{,}042 & 9{,}776 & 25{,}321 \\
Art.~13 & 457 & 40.3\% & 100.0\% & 75{,}200 & 11{,}844 & 12{,}377 & 12{,}610 & 14{,}406 & 31{,}453 \\
Art.~11 & 283 & 55.8\% & 88.7\% & 142{,}000 & 20{,}173 & 23{,}504 & 21{,}477 & 16{,}716 & 36{,}531 \\
P1-1 & 281 & 31.7\% & 38.8\% & 27{,}800 & 10{,}085 & 10{,}253 & 10{,}060 & 6{,}991 & 10{,}964 \\
Art.~10 & 263 & 47.9\% & 52.1\% & 99{,}200 & 11{,}236 & 12{,}995 & 11{,}361 & 13{,}884 & 23{,}595 \\
P7-2 & 135 & 72.6\% & 98.5\% & 146{,}900 & 13{,}733 & 18{,}022 & 15{,}432 & 17{,}317 & 37{,}200 \\
Art.~2 & 123 & 5.7\% & 25.2\% & 143{,}850 & 31{,}123 & 32{,}176 & 29{,}802 & 34{,}160 & 26{,}008 \\
Art.~14 & 77 & 27.3\% & 100.0\% & 100{,}000 & 13{,}845 & 13{,}639 & 14{,}614 & 16{,}060 & 28{,}086 \\
Art.~34 & 40 & 35.0\% & 92.5\% & 94{,}850 & 12{,}856 & 14{,}265 & 13{,}194 & 20{,}874 & 28{,}729 \\
P7-4 & 37 & 64.9\% & 73.0\% & 151{,}800 & 13{,}053 & 15{,}889 & 14{,}966 & 15{,}084 & 23{,}905 \\
P1-3 & 34 & 61.8\% & 47.1\% & 34{,}700 & 3{,}986 & 5{,}236 & 4{,}039 & 7{,}006 & 35{,}309 \\
Art.~9 & 25 & 20.0\% & 60.0\% & 126{,}375 & 37{,}488 & 34{,}802 & 38{,}606 & 32{,}586 & 33{,}206 \\
\bottomrule
\end{tabular}
\caption{Article-level side view for representative systems. Rows are
overlapping because one case can involve multiple violated Articles. MAE
and positive-award p95 are in EUR.}
\label{tab:article-side-view}
\end{table*}

\subsection{Prompt Serialisation Ablation}
\label{app:prompt-serialization-ablation}

This ablation replaces the \textsc{Facts} section with the permitted structured serialisation while holding the model and prompting regime fixed. $\Delta$ is the test MAE change relative to the corresponding raw-text input; negative values indicate lower error under serialisation. The mixed signs show that serialisation is not a uniform improvement across models or prompting regimes.

\begin{table}[H]
\centering
\small
\setlength{\tabcolsep}{6pt}
\renewcommand{\arraystretch}{1.05}
\begin{tabular}{@{}llrr@{}}
\toprule
\textbf{Prompting} & \textbf{Model} & \textbf{MAE} & \textbf{$\Delta$} \\
\midrule
Vanilla & Qwen3.5-9B & 13{,}218 & -3{,}380 \\
Vanilla & Qwen3.5-27B & 18{,}118 & -4{,}117 \\
Static CoT & Qwen3.5-9B & 11{,}815 & +1{,}033 \\
Static CoT & Qwen3.5-Plus & 16{,}602 & -2{,}478 \\
Static CoT & GPT-OSS-20B & 11{,}707 & -5{,}041 \\
Static CoT & GPT-5.4 & 14{,}384 & -1{,}882 \\
Few-shot CoT$^\dagger$ & Qwen3.5-9B & 14{,}409 & -458 \\
Few-shot CoT$^\dagger$ & GPT-OSS-20B & 12{,}074 & +400 \\
\bottomrule
\end{tabular}
\caption{Prompt serialisation ablation. $\Delta$ is MAE change relative to the corresponding text-input setting.}
\label{tab:prompt-serialization-ablation}
\end{table}

\subsection{Prediction Error Buckets}
\label{app:prediction-accuracy-buckets}

Buckets are based on relative error. For zero-award cases, a prediction is counted as exact only when the model predicts zero; any positive prediction is assigned to the $>50\%$ error bucket. The reported buckets are mutually exclusive; ``Within 10\%'' excludes exact predictions. These buckets are descriptive summaries of prediction error rather than legally defined acceptability ranges.

\begin{table*}[!t]
\centering
\small
\setlength{\tabcolsep}{4pt}
\renewcommand{\arraystretch}{0.95}
\begin{tabular}{@{}llrrrrrr@{}}
\toprule
\textbf{Model} & \textbf{Variant} & \textbf{MAE} & \textbf{$N$} &
\textbf{Exact} &
\textbf{\shortstack{Within 10\%\\err.}} &
\textbf{\shortstack{10--50\%\\err.}} &
\textbf{\shortstack{$>$50\%\\err.}} \\
\midrule
\multicolumn{8}{l}{\textit{Structured-feature trees}} \\
CatBoost & --- & 9{,}881 & 2{,}897 & 104 & 73 & 308 & 2{,}412 \\
XGBoost & --- & 10{,}117 & 2{,}897 & 65 & 63 & 286 & 2{,}483 \\
LightGBM & --- & 10{,}240 & 2{,}897 & 54 & 62 & 301 & 2{,}480 \\
\midrule
\multicolumn{8}{l}{\textit{Retrieval baselines}} \\
$k$NN & metadata & 10{,}607 & 2{,}897 & 194 & 107 & 597 & 1{,}999 \\
$k$NN & struct. feats. & 10{,}632 & 2{,}897 & 478 & 104 & 554 & 1{,}761 \\
BM25 & raw text & 13{,}390 & 2{,}897 & 507 & 89 & 428 & 1{,}873 \\
BGE-M3 & sparse & 10{,}226 & 2{,}897 & 552 & 116 & 516 & 1{,}713 \\
BGE-M3 & dense & 10{,}177 & 2{,}897 & 393 & 99 & 613 & 1{,}792 \\
\midrule
\multicolumn{8}{l}{\textit{Encoder language models}} \\
ModernBERT & raw text & 10{,}133 & 2{,}897 & 134 & 75 & 365 & 2{,}323 \\
ModernBERT & late fusion & 10{,}074 & 2{,}897 & 37 & 95 & 387 & 2{,}378 \\
Legal-Longformer & raw text & 10{,}642 & 2{,}897 & 8 & 50 & 290 & 2{,}549 \\
Legal-Longformer & late fusion & 10{,}244 & 2{,}897 & 129 & 126 & 461 & 2{,}181 \\
\midrule
\multicolumn{8}{l}{\textit{Prompted LMs}} \\
Qwen3.5-9B & zero-shot & 16{,}598 & 2{,}897 & 190 & 58 & 553 & 2{,}096 \\
Qwen3.5-27B & zero-shot & 22{,}235 & 2{,}897 & 117 & 76 & 608 & 2{,}096 \\
Qwen3.5-Plus & zero-shot & 16{,}400 & 2{,}897 & 149 & 96 & 654 & 1{,}998 \\
GPT-OSS-20B & zero-shot & 15{,}678 & 2{,}897 & 537 & 22 & 219 & 2{,}119 \\
GPT-5.4 & zero-shot & 25{,}438 & 2{,}897 & 200 & 129 & 757 & 1{,}811 \\
Qwen3.5-9B & CoT & 10{,}782 & 2{,}897 & 255 & 81 & 575 & 1{,}986 \\
Qwen3.5-27B & CoT & 12{,}905 & 2{,}897 & 113 & 113 & 692 & 1{,}979 \\
Qwen3.5-Plus & CoT & 19{,}080 & 2{,}897 & 98 & 90 & 668 & 2{,}041 \\
GPT-OSS-20B & CoT & 16{,}748 & 2{,}897 & 125 & 73 & 577 & 2{,}122 \\
GPT-5.4 & CoT & 16{,}266 & 2{,}897 & 69 & 170 & 811 & 1{,}847 \\
Qwen3.5-9B & few-shot CoT$^\dagger$ & 14{,}867 & 2{,}897 & 112 & 116 & 682 & 1{,}987 \\
Qwen3.5-Plus & few-shot CoT$^\dagger$ & 24{,}141 & 2{,}897 & 101 & 115 & 707 & 1{,}974 \\
GPT-OSS-20B & few-shot CoT$^\dagger$ & 11{,}674 & 2{,}897 & 158 & 57 & 671 & 2{,}011 \\
GPT-5.4 & few-shot CoT$^\dagger$ & 22{,}051 & 2{,}897 & 83 & 156 & 753 & 1{,}905 \\
\midrule
\multicolumn{8}{l}{\textit{Knowledge-augmented ReAct agents}} \\
Qwen3.5-Plus & Expanded & 8{,}810 & 2{,}897 & 1{,}083 & 146 & 693 & 975 \\
MiniMax-M2.7 & Expanded & 8{,}819 & 2{,}897 & 1{,}116 & 103 & 681 & 997 \\
\bottomrule
\end{tabular}
\caption{Prediction accuracy buckets on the ECtHR-NPD test pool.
Exact predictions are excluded from the Within 10\% error bucket.
The buckets are mutually exclusive and exhaustive.}
\label{tab:prediction-accuracy-buckets}
\end{table*}

 \end{document}